\documentclass{article}

\usepackage{arxiv}

{}
{}
{}

\usepackage{url}
\usepackage{multirow}
\usepackage{balance}
\usepackage{babel}
\usepackage{epstopdf}
\usepackage{epsfig}
\DeclareGraphicsExtensions{.png,.pdf}

\usepackage{lineno}

\usepackage[utf8]{inputenc} 
\usepackage[T1]{fontenc}    
\usepackage{hyperref}       
\usepackage{url}            
\usepackage{booktabs}       
\usepackage{amsfonts}       
\usepackage{nicefrac}       
\usepackage{microtype}      
\usepackage{cleveref}       
\usepackage{lipsum}         
\usepackage{graphics}
\usepackage{doi}
\usepackage{epstopdf}

\title{Advanced Image Quality Assessment for Hand- and Fingervein Biometrics}

\author{ \href{https://orcid.org/0000-0003-4836-0913}{\includegraphics[scale=0.06]{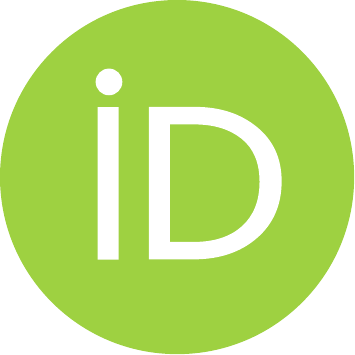}\hspace{1mm}Simon Kirchgasser}, \href{https://orcid.org/0000-0002-2716-1360}{\includegraphics[scale=0.06]{orcid.pdf}\hspace{1mm}Christof Kauba},
\href{https://orcid.org/0000-0001-5529-0154}{\includegraphics[scale=0.06]{orcid.pdf}\hspace{1mm}Georg Wimmer} and \href{https://orcid.org/0000-0002-5921-8755}{\includegraphics[scale=0.06]{orcid.pdf}\hspace{1mm}Andreas Uhl} \\
	Department of Artificial Intelligence and Human Interaction \\
	University of Salzburg \\
	Jakob-Haringer-Str. 2 \\
	5020 Salzburg, Austria \\
	\texttt{\{skirch, ckauba, gwimmer, uhl\}@cs.sbg.ac.at} \\
	
}

\renewcommand{\shorttitle}{Quality Assessment for Vasculature Biometrics}

\hypersetup{
pdftitle={Quality Assessment for Vasculature Biometrics},
pdfsubject={},
pdfauthor={Simon Kirchgasser, Christof Kauba, Georg Wimmer, Andreas Uhl},
pdfkeywords={First keyword, Second keyword, More},
}

\begin{document}
\maketitle

\begin{abstract}
Natural Scene Statistics commonly used in non-reference image quality measures and a deep learning based quality assessment approach are proposed as biometric quality indicators for vasculature images. While NIQE and BRISQUE if trained on common images with usual distortions do not work well for assessing vasculature pattern samples' quality, their variants being trained on high and low quality vasculature sample data behave as expected from a biometric quality estimator in most cases (deviations from the overall trend occur for certain datasets or feature extraction methods). The proposed deep learning based quality metric is capable of assigning the correct quality class to the vasculature pattern samples in most cases, independent of finger or hand vein patterns being assessed. The experiments were conducted on a total of 13 publicly available finger and hand vein datasets and involve three distinct template representations (two of them especially designed for vascular biometrics). The proposed (trained) quality measures are compared to a several classical quality metrics, with their achieved results underlining their promising behaviour.
\end{abstract}

\section{Introduction}\label{sec: intro}
The term vascular biometrics describes a set of biometric modalities (commonly finger as well as hand vein biometrics, but also sclera pattern based biometrics), which uniquely characterises people by their self-specific pattern of the human blood vessels system. Due to the blood vessels being located inside the human body the vessels' pattern can hardly be made visible by the use of visible light. Instead near-infrared (NIR) illumination in combination with NIR sensitive cameras are utilised during the acquisition process to render the vessels visible as dark lines in images or videos \cite{Uhl19a}. This is based on the fact that the de-oxigenated haemoglobin contained in the blood exhibiting a higher light absorption coefficient in the near-infrared spectrum than the surrounding tissue. The acquisition process of samples exhibiting vasculature patterns can be done by using either a reflected light set-up or a light transmission one. When using a reflected light set-up, the light source and the image sensor are positioned on the same side, opposite to the finger or hand that is to be captured. Thus, the NIR light emission and the recording of the reflected light are done on the same side. In the light transmission set-up, the light source and the image sensor are located on opposite sides. The finger or hand from which the vasculature pattern is to be captured is placed in between and thus, the emitted light needs to travel through the human tissue before it reaches the image sensor and can be captured. \\
The most popular vasculature biometric is finger vein recognition as the acquisition of finger vein samples is almost as easy as capturing a fingerprint, can be done in a contact-less manner and is more resistant to forging than fingerprints (a vein pattern is not left on a surface like e.g. a fingerprint is). However, especially with finger veins, the area where meaningful vessel information can be detected is limited. Depending on the acquisition set-up, this can affect the quality of the recorded biometric sample's information. In the majority of all known cases it is easier to make the vessels visible if the acquisition of the finger vein images is done from the palmar view in a light transmission set-up. Hence, most publicly available databases contain only palmar finger vein samples (see Table \ref{tab:SummaryDatasets}). In hand vein biometrics a larger area exhibiting venous patterns can be utilised to extract biometric information. 
One downside with hand vein biometrics is that usually more skin wrinkles and skin hair will be present and visible in the acquired samples than in the finger vein ones. These wrinkles and hairs tend to be mistakenly detected as blood vessels, influencing the quality of the biometric information and thus, the recognition performance of the whole biometric system. Furthermore,the different tissue layers of the human skin and the bone structure also influence the quality of the vasculature pattern samples as they absorb parts of the emitted light, which results in a reduced light intensity and thus, in a reduced contrast of the vasculature patterns and the background, lowering the quality of samples in general. \\
For the successful application of any biometric authentication system, the sample quality plays a vital role. Hence, for a vasculature based biometric authentication system, it is fundamental to determine the quality of the acquired sample data. This quality assessment is necessary for both enrolment and the actual authentication such that a re-capturing can be initiated in case the sample's quality turns out to be insufficient. Furthermore, quality assessment is useful for selecting subsequent steps in the signal processing pipeline (algorithms / parameters) and in unsupervised scenarios for user guidance during capturing. Especially, in an unsupervised scenario it is highly likely that different amounts of rotation can be detected when a finger or a hand is presented to the capturnig device. Even a small amount of finger rotation ($\pm30\textdegree$)  is known to be one of the major causes of poor recognition performance in finger vein biometrics, in particular rotation in longitudinal direction \cite{Prommegger19a}. Rotation is not only a finger vein specific quality factor, also hand vein recognition is influenced by out-of-plane hand rotations. The influence of the rotation cannot be determined based on analysis of a single sample image but only in the context of an analysis of template comparison results (rotation difference of the two samples involved in the comparison). 

\subsection{Contribution of work}
This work is an extension of our previous work "Fingervein Sample Image Quality Assessment using Natural Scene Statistics" \cite{Remy22b}.
The original idea of \cite{Remy22b} was to propose a learning-based finger vein sample quality assessment scheme, based on training natural scene statistics (NSS) on finger vein sample data as used in general purpose image quality metrics (IQM) like NIQE \cite{Mittal12} and BRISQUE \cite{Mittal12a}. In order to ensure sufficient generalisability (i.e. being independent from the utilised recognition features), the IQMs are trained on low and high quality sample data, classified by human judgement, respectively. So overall, the authors introduced (i) a learning based scheme, which uses (ii) NSS, a well established statistical model to capture image quality variations and employed (iii) training-data based on human scoring to provide independence from specific features used in particular recognition schemes. Hence, the aim was to demonstrate a sufficient generalisability among different datasets and finger vein recognition schemes, respectively. \\
While in \cite{Remy22b} the evaluation of the proposed quality assessment scheme was only done on two finger vein databases, this work evaluates the methodology on $9$ finger vein databases. Furthermore, it is extended by including another modality, hand vein biometrics, represented by additional $4$ databases. Hence, a total of $13$ vasculature databases are evaluated by using a broad set of various quality assessment methods. The quality schemes based on NSS have been re-trained to include characteristics of the additional databases instead of applying the re-trained versions of BRISQUE and NIQE as proposed in \cite{Remy22b}. Due to the fact that in recent years the usage of deep-learning (DL) based applications for quality evaluation became feasible as well (see Section \ref{sec: relatedWork}), a new DL based method for vascular quality assessment is proposed in this research article as well. \\
The remainder of the paper is organised as follows: Section \ref{sec: relatedWork} provides an overview on related work in quality assessment for vascular biometrics. Section \ref{sec: datasets} describes the utilised datasets, while the applied training based non-reference quality metrics are described in \ref{sec: nss} and the proposed DL based quality estimation method in \ref{sec: DLQMethod}. Subsequently, Section \ref{sec: expSetup} states the details about the experimental protocol, whereas the experimental results are presented and discussed Section \ref{sec: evaluation}. Finally, the work is concluded in Section \ref{sec: conclusion}.

\section{Related Work on Biometric Quality Evaluation}\label{sec: relatedWork}
Biometric sample quality measures are applied in order to estimate if a recorded sample of a specific biometric trait can successfully be evaluated by an automated biometric recognition system. Hence, the ISO/IEC 29794:2016 Biometric Sample Quality standard contains a definition on how quality evaluation can be performed for most biometric modalities. As a consequence, for the well-established and widely employed biometric modalities like fingerprint, face or iris dedicated image quality evaluation algorithms have been established and successfully applied, e.g. \cite{Grother07,Bharadwaj14}. However, as already discussed in our previous work \cite{Remy22b}, the ISO/IEC 29794:2016 Biometric Sample Quality standard does not yet include a unified quality evaluation criterion for vasculature sample images. \\
Several studies on finger and hand vein quality evaluation were published in past 10 years. Based on these publications, two main classes vasculature quality assessment techniques can be distinguised: \textit{(a)} non-vasculature-feature based techniques and \textit{(b)} vasculature-features-based ones. Methods belonging to the first class can directly be used on finger or hand vein samples after the acquisition process is completed, while methods belonging to the second class need to extract vascular specific features prior to the application of the quality measure. In the following a short description of methods from these two classes is given. \\
Techniques from the first group make use of low-level image information like {\it gradient}, {\it contrast}, {\it entropy}, {\it clarity} and {\it brightness} uniformity. Methods utilising {\it gradient}, {\it contrast} and {\it entropy} information have been proposed in \cite{Yang13} and \cite{Matkovic05} for an application on finger vein samples. The latter named three low-level features can also be used in a combined manner as proposed by \cite{Peng14}. The combination of {\it gradient}, {\it contrast} and {\it entropy} can be done by fusing them using a {\it triangular} norm scheme. In \cite{Peng14} the {\it triangular} norm scheme was also applied to finger vein images, while in \cite{Wang17} {\it clarity} and {\it brightness} uniformity information was used to estimate the quality of palmvein samples. \\
Other methods using non-vasculature specific features are based either on {\it Radon} transform as e.g. in \cite{Qin17} or on the use of the noise ratio based on human visual system (HSNR) method \cite{Ma13}. HSNR tries to represent the human visual system to evaluate the quality of vasculature samples by combining four different indices: the image contrast, the deviation of foreground areas’ centre of mass with respect to the geometric centre of the whole image, the effective area of an image, the area and locations where vascular information can be detected and the signal to noise index adapted with respect to the human visual system. \\
Techniques belonging to the second group make use of vasculature specific features, that need to be extracted during or prior to the quality evaluation process. In \cite{Nguyen13} the same features that are subsequently used during the recognition process are also utilised to estimate the biometric sample quality. The authors used the number of pixels representing vascular pattern information as the main feature for the quality estimation. In several studies employing learning-based approaches (e.g. \cite{Quin17,Quin19,Ren22}) incorrect and/or poor template comparisons in recognition experiments are analysed and the gained information is used to improve the learning procedure. Furthermore, a traditional CNN can be trained to establish a vascular quality measurement as described in \cite{Zeng18}. Most recent in our previous work \cite{Remy22b} a learning-based vascular sample quality assessment scheme was proposed which is based on re-training the general purpose image quality metrics NIQE \cite{Mittal12} and BRISQUE \cite{Mittal12a} on vascular images. This study found that due to the high differences between vascular sample images and typical natural scene images a re-training of the quality estimators is mandatory. Similar to the original training of NIQE and BRISQUE, the re-training of NIQE is done by using only high quality images, while for BRISQUE high quality images and low quality images are considered. The results indicated that the re-trained versions of both measures provide better results on the finger vein samples than compared to their original versions. \\
As mentioned in the introduction of this study (see Section \ref{sec: intro}), the findings of \cite{Remy22b} are extended/validated by including additional vascular databases (especially also including hand vein ones as a further biometric trait) and feature types as well as proposing a new deep learning based method for vascular quality estimation (see Section \ref{sec: DLQMethod}). Hence, most of the quality measures evaluated in \cite{Remy22b} are included in this study as well. This includes the following one from the first class (non-vasculature feature based): global contrast factor (GCF) \cite{Matkovic05}, grey level energy score (EntropyBased) (based on the entropy information of the image \cite{Yang13}), the Radon transformation approach (Radon) \cite{Qin17}, the triangular norm scheme (TNorm) \cite{Peng14} and the scheme proposed in \cite{Wang17} (Wang). Furthermore, HSNR \cite{Ma13} is also included. From the second group (vasculature feature based) only the re-trained versions of NIQE and BRISQUE are used, but not the same ones as in \cite{Remy22b}. Instead new versions which have been trained on a larger number of training data are utilised. 

\section{Datasets}\label{sec: datasets}
\begin{figure*}[!t]
\centering
\caption{Example images of the vascular finger images utilised as reference datasets. \label{fig: FVimpressions}}
{\begin{tabular*}{\textwidth}{@{\extracolsep{\fill}}ccc}
\includegraphics[scale=0.4]{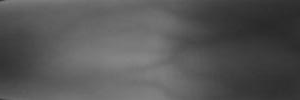} & \includegraphics[scale=0.4]{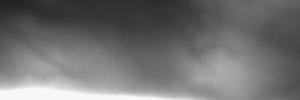} & \includegraphics[scale=0.4]{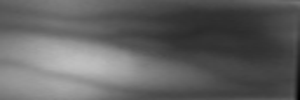} \\

FV-USM~\cite{BAsaari14a} & HKPU-FV~\cite{Kumar11} & MMCBNU\_6000~\cite{BLu13a} \\
 
\includegraphics[scale=0.4]{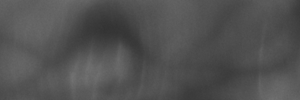} & \includegraphics[scale=0.4]{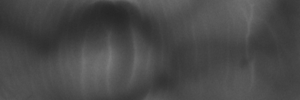} & \includegraphics[scale=0.4]{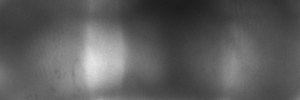}\\

PLasDOR~\cite{Kauba18d} & PLEDDOR~\cite{Kauba18d} & PLasPAL~\cite{Kauba18d} \\

\includegraphics[scale=0.4]{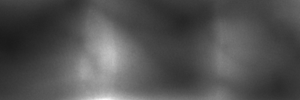} & \includegraphics[scale=0.4]{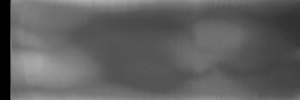} & \includegraphics[scale=0.4]{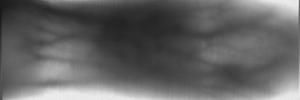}\\

PLEDPAL~\cite{Kauba18d} & SDUMLA~\cite{Yin11a} & UTFVP~\cite{BTon13a} \\
\end{tabular*}}{}
\end{figure*}

\begin{figure*}[!t]
\centering
\caption{Example images of the vascular hand images utilised as reference datasets. \label{fig: HVimpressions}}
{\begin{tabular*}{\textwidth}{@{\extracolsep{\fill}}cccc}
\includegraphics[scale=0.34]{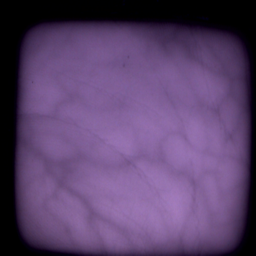} &  \includegraphics[scale=0.34]{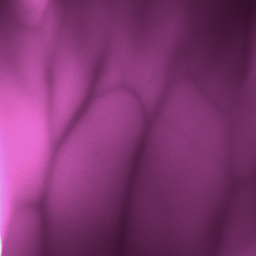} &
\includegraphics[scale=0.34]{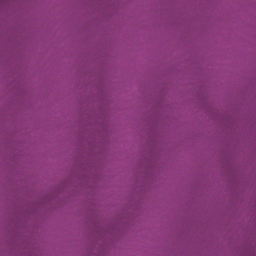} &  \includegraphics[scale=0.34]{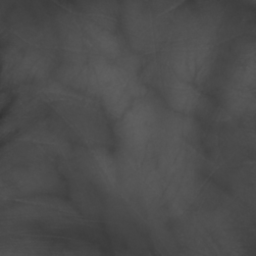}\\

CIE-HV~\cite{Kabacinski11a} & PTrans~\cite{Sequeira18a}  & PRefl~\cite{Sequeira18a} & VERA~\cite{Tome15a}\\
\end{tabular*}}{}
\end{figure*}

The experiments evaluating the discussed vasculature specific quality metrics were conducted on $13$ publicly available vascular pattern databases  (see Table \ref{tab:SummaryDatasets}). Four of these databases are hand vein ones, while the remaining ones contain finger vein images captured from palmar or dorsal view. Example impressions of the utilised databases are visualised in Figure \ref{fig: FVimpressions} and Figure \ref{fig: HVimpressions}.

\begin{table}[ht!b] 
	\centering
	\caption{This table contains the key information of the utilised vasculature databases: the database name, the capturing perspective (dorsal or palmar: dor/pal), the number of subjects (subj), the number of fingers per subject (fgs), the total number of images (imgs), the number of sessions (s) and the corresponding image size. \label{tab:SummaryDatasets}}
	{\begin{tabular*}{25pc}{@{\extracolsep{\fill}}ccccccc@{}}\hline 
		database name & dor/pal & subj & fgs & imgs & s & image size\\
		\hline
		\textbf{finger vein} & & & & & &\\ 
		\hline
		FV-USM~\cite{BAsaari14a} & palmar & 123 & 4 & 5940 & 2 & 640$\times$480\\
		HKPU-FV~\cite{Kumar11} & palmar & 156 & 6 & 6264 & 2 & 513$\times$256 \\
		MMCBNU\_6000~\cite{BLu13a} & palmar & 100 & 6 & 6000 & 1 & 640$\times$480\\
		PLasDOR~\cite{Kauba18d} & dorsal & 60 & 6 & 3600 & 1 & 1280$\times$1024\\
		PLEDDOR~\cite{Kauba18d} & dorsal & 60 & 6 & 3600 & 1 & 1280$\times$1024\\
		PLasPAL~\cite{Kauba18d} & palmar & 60 & 6 & 3600 & 1 & 1280$\times$1024\\
		PLEDPAL~\cite{Kauba18d} & palmar & 60 & 6 & 3600 & 1 & 1280$\times$1024\\
		SDUMLA~\cite{Yin11a} & palmar & 106 & 6 & 3816 & 1 & 320$\times$240\\
		UTFVP~\cite{BTon13a} & palmar & 60 & 6 & 1440 & 2 & 672$\times$380\\
		\hline
		\textbf{hand vein} & & & & & &\\ 
		\hline 
		CIE-HV \cite{Kabacinski11a} & palmar & 50 & 8 & 1200 & 3 & 1280$\times$960 \\
		PTrans\cite{Sequeira18a} & dorsal & 40 & 5 & 400 & 1 & 384$\times$384\\
		PRefl\cite{Sequeira18a} & dorsal & 40 & 5 & 400 & 1 & 384$\times$384\\
		VERA \cite{Tome15a} & palmar & 110 & 5 & 2200 & 2 & 580$\times$680\\
		\hline
	\end{tabular*}}{}
\end{table}

The \textbf{Finger Vein USM} (\textbf{FV-USM}) \cite{BAsaari14a} contains 5904 palmar finger vein images with an image resolution of $640 \times 480$ pixels. The images have been acquired from 123 subjects in 2 independent sessions. The image capturing process was the same for all subjects and both sessions, resulting in 6 images per finger, whereby a total of 4 fingers were recorded for each subject. \\
The \textbf{Hong Kong Polytechnic University Finger Image Database (Version 1.0)} (\textbf{HKPU-FV}) \cite{Kumar11} is composed by 6264 palmar vascular finger images, exhibiting a resolution of 513 $\times$ 256 pixels. These images have been acquired from 156 volunteers in 2 sessions. 6 image samples each are recorded from the index and middle finger of the left hand. \\
The \textbf{Chonbuk National University MMCBNU-6000 finger vein database} (\textbf{MMCBNU\_6000}) \cite{BLu13a} contains 6000 palmar light transmission finger vein images (resolution of $640 \times 480$ pixel) acquired from 100 subjects. For each subject 6 fingers (10 images per finger) have been acquired in a single session. \\
The \textbf{PLUSVein-FV3 Finger Vein Data Set} (\textbf{PLUSVein-FV3}) \cite{Kauba18d} is composed by a total of 4 subsets using two different capturing devices both being capable of acquiring samples from the palmar as well as the dorsal view. In Table \ref{tab:SummaryDatasets} these subsets are mentioned separately as \textit{PLasDOR}, \textit{PLEDDOR}, \textit{PLasPAL} and \textit{PLEDPAL}. Each subset contains 1800 images, the samples having an image resolution of $200 \times 750$ pixels, acquired from 60 subjects in a single session. In total 6 fingers per subject (index, middle and ring finger of both hands) and 5 images per finger have been recorded. \\
The \textbf{Shandong University Machine Learning and Applications - Homologous Multi-modal Traits Database} (\textbf{SDUMLA-HMT}) \cite{BYin11a} contains several biometric modalities (as the name implies). In the current study only the subset containting the vascular finger patterns is used. This subset contains a total of 3816 palmar image from 106 subjects (each image has a resolution of 320 $\times$ 240 pixels). From each subject the 6 finger vein samples of the index, the middle and the ring finger of both hands were captured. \\
The \textbf{University of Twente Finger Vascular Pattern Database} (\textbf{UTFVP}) \cite{BTon13a} contains 1440 palmar finger vein sample images, acquired in two sessions from 60 different subjects. For each subject vascular pattern images of the ring, middle and index finger from both hands have been captured (two samples per finger/session). The acquired images exhibit a resolution of 672 $\times$ 380 pixels.

The \textbf{University of Poznan Hand Vein Data Set} (\textbf{CIE-HV}) \cite{Kabacinski11a} contains 1200 hand vein images, which have been acquired from the palmar view in a reflected light illumination set-up, exhibiting a resolution of $1280 \times 960$ pixels. Images of both hands (4 samples per hand) from 50 subjects were acquired in 3 sessions. \\
The \textbf{PROTECT Hand Vein Dataset} (\textbf{PROTECT-HV}) \cite{Sequeira18a} is composed of 2 subsets. For the acquisition of each subset a different illumination technique (light transmission or reflected) was used. Hence, in Table \ref{tab:SummaryDatasets} these subsets are mentioned separately as \textit{PTrans}, \textit{PRefl} In contrast to the CIE-HV, the samples have not been captured from the palmar, but the dorsal view. Each subset contains a total of 400 images from both hands of 40 subjects (5 images per hand). All images exhibit a resolution of $384 \times 384$ pixels. \\
The \textbf{Idiap Research Institute VERA Fingervein Database} (\textbf{VERA}) \cite{Tome15a} contains 2200 palmar hand vein images, recorded from 110 subjects in 2 sessions from both hands using reflected light illumination. The images exhibit a resolution of $580 \times 680$ pixels. 

\section{Natural Scene Statistics in Vascular Image Quality} \label{sec: nss}
In the following details about non-reference image quality metrics, particularly focussing on methods based on the concept of natural scene statistics, are described.

\subsection{Non-reference image quality metrics}
Current state-of-the-art non-reference Image Quality Assessment (NR IQM) algorithms are based on models that are able to learn to predict human judgments from databases of human-rated, distorted images. These kinds of IQM models are necessarily limited, since they can only assess quality degradations arising from the distortion types that they have previously seen and been trained on. However, it is also possible to contemplate sub-categories of general-purpose NR IQM models having tighter conditions. A model is said to be opinion-aware (OA) if it has been trained on a database(s) of human rated distorted images and associated subjective opinion scores. Algorithms like BRISQUE as described below are OA IQM measures. However, IQM like NIQE (see below), are opinion-unaware (OU) and they make only use of measurable deviations from statistical regularities observed in natural images {\bf without being trained on human-rated distorted images} and indeed without any exposure to distorted images. \\
Systematic comparisons of the NR IQM as used in this paper have been published in \cite{Nouri13,Charrier15}. Both, in non-trained \cite{Nouri13} as well as in specifically trained manner \cite{Charrier15}, the correspondence to human vision turns out to be highly dependent on the dataset considered and the type of distortion present in the data. Thus, no ``winner'' has been identified among the techniques considered with respect to correspondence to subjective human judgement and objective distortion strength.

\subsection{BRISQUE - Blind/Referenceless Image Spatial Quality Evaluator}
BRISQUE \cite{Mittal12} is a natural scene statistic based spatial NR QA algorithm. It is based on the principle that natural images possess certain regular statistical properties that are measurably modified in the presence of distortions. \\
In the first stage an image is locally normalised (via local mean subtraction and divisive normalisation), resulting in the so called \textit{mean subtracted contrast normalized} (MSCN) coefficients. An AGGD (Asymmetric Generalized Gaussian Model) distribution is used to fit the MSCN statistic from pristine as well as distorted images - in order to quantify the dependency between neighbours, the relationships between adjacent MSCN coefficients are analysed via pairwise products at a distance of 1 pixel along four orientations. The parameters $(\mu, \nu, \sigma^2_l, \sigma^2_r) $ of the best AGGD fit are extracted for each orientation, which leads to a total of 16 parameters (4 parameters/orientation $\times$ 4 orientations). Because images are naturally multi-scale, and distortions affect their structure across scales, these features are extracted at two scales - the original image scale, and at a reduced resolution. Thus, a total of 32 features are selected - 16 at each scale. \\
A set of pristine images from the Berkeley image segmentation database is taken, additionally similar types of distortions as present in the LIVE image quality database were introduced in each image at varying degrees of severity to form the distorted image set: JPEG 2000, JPEG, white noise, Gaussian blur, and fast fading channel errors (thus, BRISQUE is a OA IQM). The computed AGGD features in combination with their associated difference mean opinion score are used to train a probabilistic support vector regression model (SVR), which is then used for classification. The difference mean opinion score represents the subjective quality of each image and is obtained by averaging across humans for each of the visual signals in the study.

\subsection{NIQE - Natural Image Quality Evaluator}
NIQE \cite{Mittal12a} is a NR OU-DU IQM (no reference, opinion unaware \& distortion unaware). Thus, it uses only measurable deviations from statistical regularities in natural images, without training on human-rated distorted images. The NSS features used in the NIQE index are similar to those used in BRISQUE, however, NIQE only uses the NSS features of natural images and is not - as BRISQUE - trained on features obtained from both natural and distorted images (and the corresponding human judgments of the quality of the latter). As a consequence, the NIQE index is not tied to any specific distortion type, while BRISQUE is limited to the types of distortions it has been trained on and tuned to. The MSCN coefficients are computed in P$\times$P  image patches, but only patches with sufficient sharpness are selected for further processing. NIQE is applied by computing the 32 identical NSS features from those patches fitting them with the AGGD model, and then comparing its fit to the AGGD model derived from pristine images.

\section{Vascular Quality estimation using DL}\label{sec: DLQMethod}
Opposed to the quality methods discussed in Section \ref{sec: relatedWork}, the newly designed metric to predict the quality of finger and hand vein images is a CNN-based one. For training the CNN, the biometric image data discussed in the previous Section \ref{sec: datasets}, which is separated manually into the quality classes poor, middle and good quality, is used. The detailed procedure of the class separation and further details regarding the experimental protocol are given in the subsequent Section \ref{sec: expSetup}. The CNN is based on the Squeeze-Net (SqNet) \cite{Moskewicz16} architecture, whereby the employed SqNet has already been pre-trained on the ImageNet database (\url{http://www.image-net.org/}). The finger and hand vein images are utilised to refine the pre-trained CNN, such that the net is more sensitive to this specific type of biometric data. \\
As loss function for the refined CNN training the triplet loss function \cite{Schroff15} is applied. The triplet loss requires three input images at once (a so called triplet) as CNN input, where two images are from the same quality level (the so called \textit{Anchor} and another image with the same quality level, further denoted as \textit{Positive}) and the third image is from a different quality level (further denoted as \textit{Negative}). Note, that it is not of importance if the three images contain the biometric information of the same person, indeed they are randomly selected from the assigned quality classes. Using the triplet loss, the network learns to minimize the distance between the \textit{Anchor} and the \textit{Positive} while maximising the distance between the \textit{Anchor} and the \textit{Negative}. The triplet loss using the squared Euclidean distance (as used in the study) is defined as follows:

\begin{equation}
    L(A,P,N)=     \max(E_P - E_N +m, 0), \nonumber
    \label{eq:defTripletLoss}
\end{equation}
where $A$ is the \textit{Anchor}, $P$ the \textit{Positive} and $N$ the \textit{Negative}. $m$ is a margin that is enforced between positive and negative pairs and is set to $m=1$. $f(I)$ is an embedding (the CNN output) of an input image $I$. $E_P = ||f(A)-f(P)||^{2}$ and $E_N = ||f(A)-f(N)||^{2}$ denote the squared Euclidean distance between the CNN outputs of the \textit{Anchor} and the \textit{Positive} as well as the \textit{Negative}, respectively.  \\
A triplet of training images (\textit{Anchor}, \textit{Positive} and \textit{Negative}) is processed through the CNN resulting in a CNN output for each of the three images. This CNN outputs are then used to compute the triplet loss to update the CNN. In order to only select triplets for training that are able to improve the model, we employ hard triplet selection \cite{Schroff15}. Hard triplet selection only permits those triplets ($A,P,N$) for training with $L(A,P,N)>0$. \\
Summarised this means the CNN is trained to create a CNN output $f(I)$, such that the squared Euclidean distances between CNN outputs of images from same quality levels are small, whereas the Euclidean distances between CNN outputs of any pairs of images from different quality levels are large. So basically, the CNN clusters images of the same quality level together in the CNN output space, apart from images of other quality levels. \\
However, the applied CNN architecture is not able to directly predict the quality level of an image. For this prediction task an SVM is applied additionally. The SVM is trained using the CNN outputs from the training data. To predict the quality level of an evaluation image, the image is fed to the CNN and then the SVM classifier predicts the quality level based on the CNN output.

\section{Experimental Protocol}\label{sec: expSetup}
\begin{figure*}[!t]
\centering
\caption{Example images of vascular finger hand hand images categorised into the manually selected quality levels \textit{poor}, \textit{middle} and \textit{good}. \label{fig: qualityClasses}}
{\begin{tabular*}{\textwidth}{@{\extracolsep{\fill}}ccc}
\textit{poor} & \textit{middle} & \textit{good} \\
\includegraphics[scale=0.4]{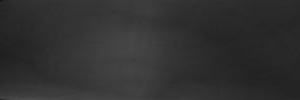} & \includegraphics[scale=0.4]{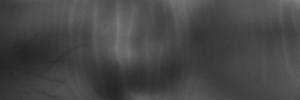} & \includegraphics[scale=0.4]{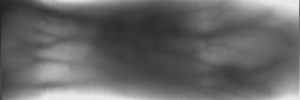} \\
\includegraphics[scale=0.4]{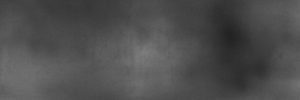} & \includegraphics[scale=0.4]{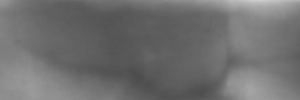} & \includegraphics[scale=0.4]{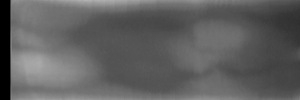}\\
\includegraphics[scale=0.4]{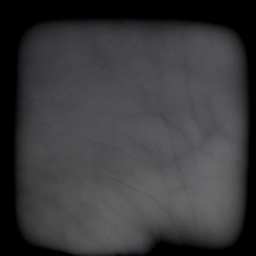} & \includegraphics[scale=0.4]{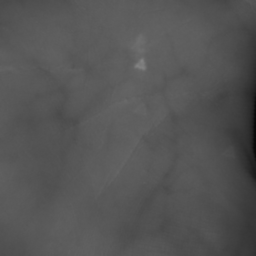} & \includegraphics[scale=0.4]{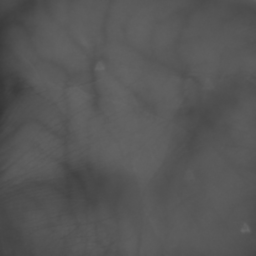}\\
\end{tabular*}}{}
\end{figure*}

Finger detection, finger alignment and RoI extraction (being mandatory for the stable application of quality metrics) for all finger and hand vein databases is done as described in \cite{BLu13b}. After pre-processing, the extracted features are used to perform the baseline experiments, resulting in the equal error rate (EER), the false match rate for a false non-match rate less or equal to $0.1\%$ (FMR1000) and zero false match rate (ZeroFMR) for each of the aforementioned databases. The experiments are conducted by utilising the PLUS OpenVein Finger- and Hand-Vein Toolkit (\url{http://www.wavelab.at/sources/OpenVein-Toolkit/} \cite{Kauba19b}). \\
To extract the feature information from the given vasculature patterns three very distinct techniques are applied: feature extraction schemes that are binary vessel structure based (i) \textit{Gabor Filter (GF)} \cite{BKumar12a}, (ii) \textit{Maximum Curvature (MC)} \cite{BMiura07a} and (iii) keypoint based \textit{SIFT} \cite{Kauba14a}. The GF and MC feature templates are subsequently compared using a correlation-based approach proposed in \cite{BMiura07a}, the so called Miura matcher, while SIFT-based recognition is applied as described in \cite{Kauba14a}. \\
Prior to the training of NIQE, BRISQUE and the DL method and the quality metric evaluation each of the databases has been manually assigned to three quality levels: \textit{poor}, \textit{middle} and \textit{good}. There are two reasons for this separation of the samples: First, as this study is an extension of \cite{Remy22b}, the quality metrics BRISQUE and NIQE needed to be retrained using more data. Due to the design of those metrics a separation into different quality classes is mandatory, otherwise the applied training of the natural scene statistics based linear support vector machine (SVM) classifier would not be possible. Second, the new deep-learning (DL) based quality metric is designed to separate different classes of quality levels. Hence, a class specific data separation is necessary for training this method. Note, it is possible that images originally recorded from the same subject may be selected in different classes. Table \ref{tab: DBsplit} lists the number of images belonging to one of the three classes for each database including the average quality values per class of HSNR and Wang. Example images for each of the quality classes are presented in Figure \ref{fig: qualityClasses}.

\begin{table}[ht!b] 
	\centering
	\caption{Overview on the number of images contained in each of the three manually selected classes \textit{poor}, \textit{middle} and \textit{good}. For each database and class the HSNR and Wang quality values are shown as well. \label{tab: DBsplit}}
	{\begin{tabular*}{26pc}{@{\extracolsep{\fill}}ccccc@{}}\hline 
		\multirow{2}{*}{database name} & nr. of images/ & \multirow{2}{*}{\textit{poor}} & \multirow{2}{*}{\textit{middle}} & \multirow{2}{*}{\textit{good}}\\
		& quality methods & & & \\
		\hline
		\textbf{finger vein - dorsal} & & 936 & 817 & 1790 \\ 
		\hline
		& \textit{nr. img} & 520 & 385 & 895 \\
		PLasDOR & \textit{HSNR} & 59.84 & 59.84 & 59.97 \\ \vspace{0.5em}
		& \textit{Wang} & 0.24 & 0.23 & 0.21 \\ 
		& \textit{nr. img} & 416 & 489 & 895 \\
		PLEDDOR & \textit{HSNR} & 59.72 & 59.70 & 59.91 \\
		& \textit{Wang} & 0.23 & 0.22 & 0.19 \\
		\hline
		\textbf{finger vein - palmar} & & 1977 & 11747 & 10167\\ 
		\hline
		& \textit{nr. img} & 235 & 3723 & 1946 \\
		FV-USM & \textit{HSNR} & 59.27 & 59.25 & 59.84 \\ \vspace{0.5em}
		& \textit{Wang} & 0.13 & 0.13 & 0.14 \\
		& \textit{nr. img} & 495 & 2009 & 628 \\
		HKPU-FV  & \textit{HSNR} & 60.95 & 59.99 & 60.20 \\ \vspace{0.5em}
		& \textit{Wang} & 0.14 & 0.11 & 0.11 \\
		& \textit{nr. img} & 346 & 3415 & 2239 \\
		MMCBNU & \textit{HSNR} & 61.04 & 60.99 & 60.98 \\ \vspace{0.5em}
		& \textit{Wang} & 0.08 & 0.08 & 0.10 \\
		& \textit{nr. img} & 305 & 792 & 703 \\
		PLasPAL & \textit{HSNR} & 59.73 & 59.75 & 59.81 \\ \vspace{0.5em}
		& \textit{Wang} & 0.19 & 0.18 & 0.18 \\
		& \textit{nr. img} & 164 & 776 & 859 \\
		PLEDPAL & \textit{HSNR} & 59.55 & 59.80 & 59.80 \\ \vspace{0.5em}
		& \textit{Wang} & 0.20 & 0.18 & 0.19 \\
		& \textit{nr. img} & 308 & 804 & 2704 \\
		SDUMLA & \textit{HSNR} & 60.83 & 59.56 & 59.49 \\ \vspace{0.5em}
		& \textit{Wang} & 0.15 & 0.16 & 0.15 \\
		& \textit{nr. img} & 124 & 228 & 1088 \\
		UTFVP & \textit{HSNR} & 60.44 & 60.52 & 60.18 \\ \vspace{0.5em}
		& \textit{Wang} & 0.13 & 0.12 & 0.13 \\
		\hline
		\textbf{hand vein} & & 749 & 1332 & 1360\\ 
		\hline 
		& \textit{nr. img} & 80 & 436 & 684  \\
		CIE-HV & \textit{HSNR} & 60.67 & 60.67 & 60.62 \\ \vspace{0.5em}
		& \textit{Wang} & 0.25 & 0.25 & 0.25 \\
		& \textit{nr. img} & 134 & 233 & 270 \\
		PTrans & \textit{HSNR} & 57.08 & 56.91 & 58.55 \\ \vspace{0.5em}
		& \textit{Wang} & 0.26 & 0.24 & 0.20 \\
		& \textit{nr. img} & 132 & 152 & 320 \\		
		PRefl & \textit{HSNR} & 59.32 & 59.67 & 58.74 \\ \vspace{0.5em}
		& \textit{Wang} & 0.57 & 0.60 & 0.57 \\
		& \textit{nr. img} & 403 & 511 & 86 \\
		VERA & \textit{HSNR} & 59.45 & 59.00 & 60.52 \\ \vspace{0.5em}
		& \textit{Wang} & 0.32 & 0.33 & 0.33 \\
		\hline
	\end{tabular*}}{}
\end{table}

Percentage-wise, there are far more images categorised into the poor class originating from the hand vein and dorsal finger vein databases compared to palmar finger vein ones. This is not only due to the objective assessment of the persons involved in the manual selection process, but also due to the fact that in dorsal images (especially in the finger vein ones) skin folds are often more prominent and visible than venous structures. Hence, the subjective perception of existing venous structures will be reduced, since a human observer focusses more on the skin folds during the quality assessment process or these folds simply overlay the existing venous structures. If the subjective classification is correct for most images, a clear quality difference between the classes middle and good compared to poor will be measurable. A large difference between middle and good would be desirable, but since the main subjective difference between middle and good was the clarity of vein structure (contrast between the vein lines and the background), it is likely that the metrics compensate for any nuance-based differences and thus, reflect similar good quality in these two groups, i.e. they quality scores of both groups will not be significantly different. The reported quality values for HSNR and Wang (see Table \ref{tab: DBsplit}) are clearly not reflecting the desired behaviour. Instead, the reported values are almost similar across all manually selected quality classes or even tend to decrease with high subjective quality. This overall trend is the same for the other, un-trained quality assessment methods, clearly indicating that those are not suitable to reflect the chosen quality classes. As it has been shown in \cite{Remy22b} a re-training of the NSS based metrics on the vasculature databases improved the relation between the assessed quality values and the subjective quality. This motivated the refined re-training of BRISQUE and NIQE done in this work. \\

For the training of BRISQUE and NIQE two different experimental protocols are chosen: (i) leave one dataset out and (ii) 10-fold sampling. In the first case all \textit{good} (NIQE) or \textit{poor} and \textit{good} (BRISQUE) finger or hand vein images are chosen as training set, but excluding one particular databases' images e.g. MMCBNU, if MMCBNU is the one it shall be evaluated later on. Thus, the images and corresponding characteristics of one database are never included in the training process. The second protocol is used to validate the stability of the re-trained versions of BRISQUE and NIQE and the independence from the quality values from the particular vasculature pattern samples (each random subset of the same class should exhibit the same average quality value). A traditional randomly performed 10-fold sampling is conducted. The images of all datasets from the same image type are combined together and then divided into ten folds, where each fold consists of a tenth of the subjects from the combined data. It does not matter from which dataset the images originate but all images of the same subject have to be in the same fold. This is done to prevent any bias between the training and evaluation data. Images that have been selected to be used as training images are excluded to be part of the evaluation databases and hence, are excluded from subsequently performed evaluations. \\
In case of the training for DL method the experiments and training procedures are applied separately for the three types of samples/biometric traits that are evaluated (palmar FV, dorsal FV, HV) using a 4-fold cross validation, similar to the 10-fold cross-validation protocol as described before. Each fold is applied once for evaluation and the images of the remaining three folds are used as training data. The entire DL-based method is trained on the training portion and subsequently evaluated on the evaluation fold. \\
For calculating the quality scores the IQM described in Section \ref{sec: nss} and the DL-based method (see Section \ref{sec: DLQMethod}) are employed. While the DL-based method is implemented in Python, all other metrics ({\it EntropyBased}, {\it GCF}, {\it HSNR}, {\it Radon}, {\it TNorm} and {\it Wang17}) are implemented in MATLAB. For BRISQUE and NIQE the MATLAB implementations from the developers of NIQE and BRISQUE (all available from \url{http://live.ece.utexas.edu/research/quality/}) are utilised. In all cases for the latter metrics (i) the default settings and (ii) trained with the vasculature data as described are evaluated. For these two applied metrics lower values indicate better quality, while for all other metrics higher values indicate better quality. \\
In the previous work \cite{Remy22b} three experiments were conducted. In this extension only the third one is repeated: The third experiment successively discards an increasing number of low quality sample images from the datasets (sorted according to quality, lowest quality discarded first) and compares the verification EER of these datasets exhibiting increasingly higher quality with the EER of the original ones. Hence, the third experiment assesses if the quality measures actually serve their purpose: Does the recognition accuracy improve if low quality samples are filtered out? \\
Additionally the single databases are thoroughly evaluated for the first time using this selective variety of different methodologies to assess the vasculature samples' quality. These experiments are represented by a statistical evaluation.

\begin{figure*}[!b]
	\begin{minipage}{0.33\textwidth}
	\centering{\includegraphics[width=1\textwidth]{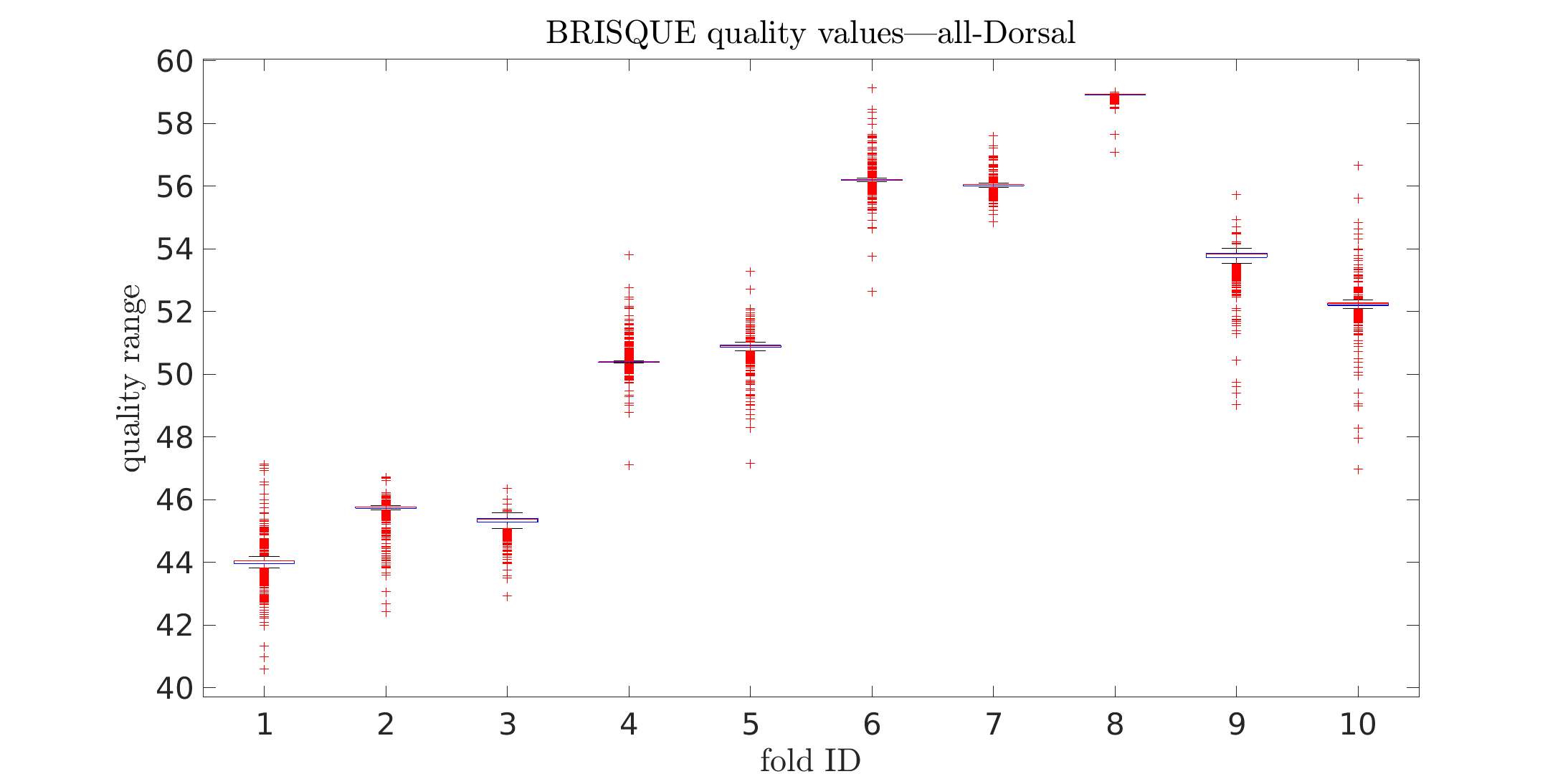}}
	\end{minipage}
	\begin{minipage}{0.33\textwidth}
	\centering{\includegraphics[width=1\textwidth]{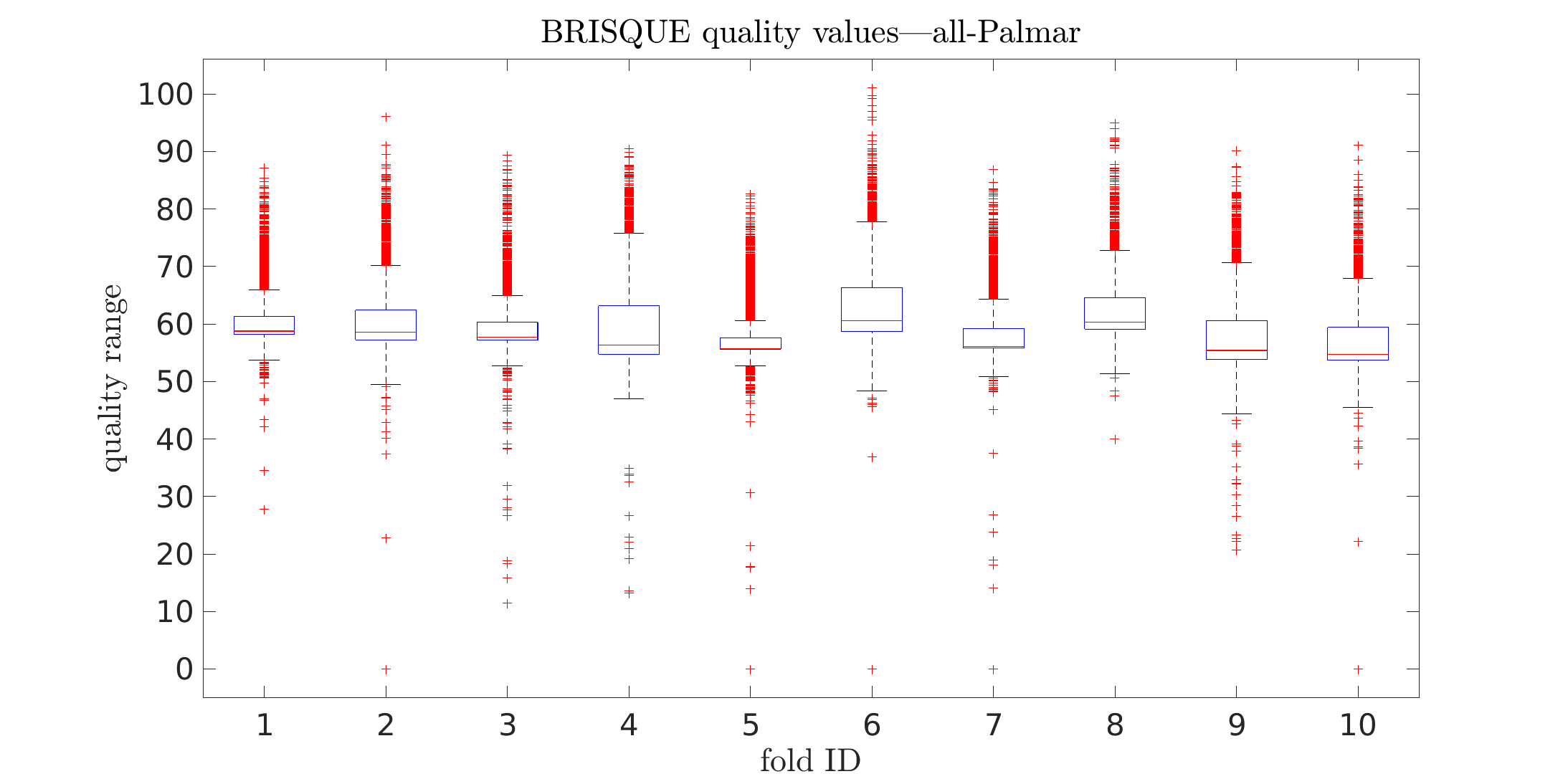}}
	\end{minipage}
	\begin{minipage}{0.33\textwidth}
	\centering{\includegraphics[width=1\textwidth]{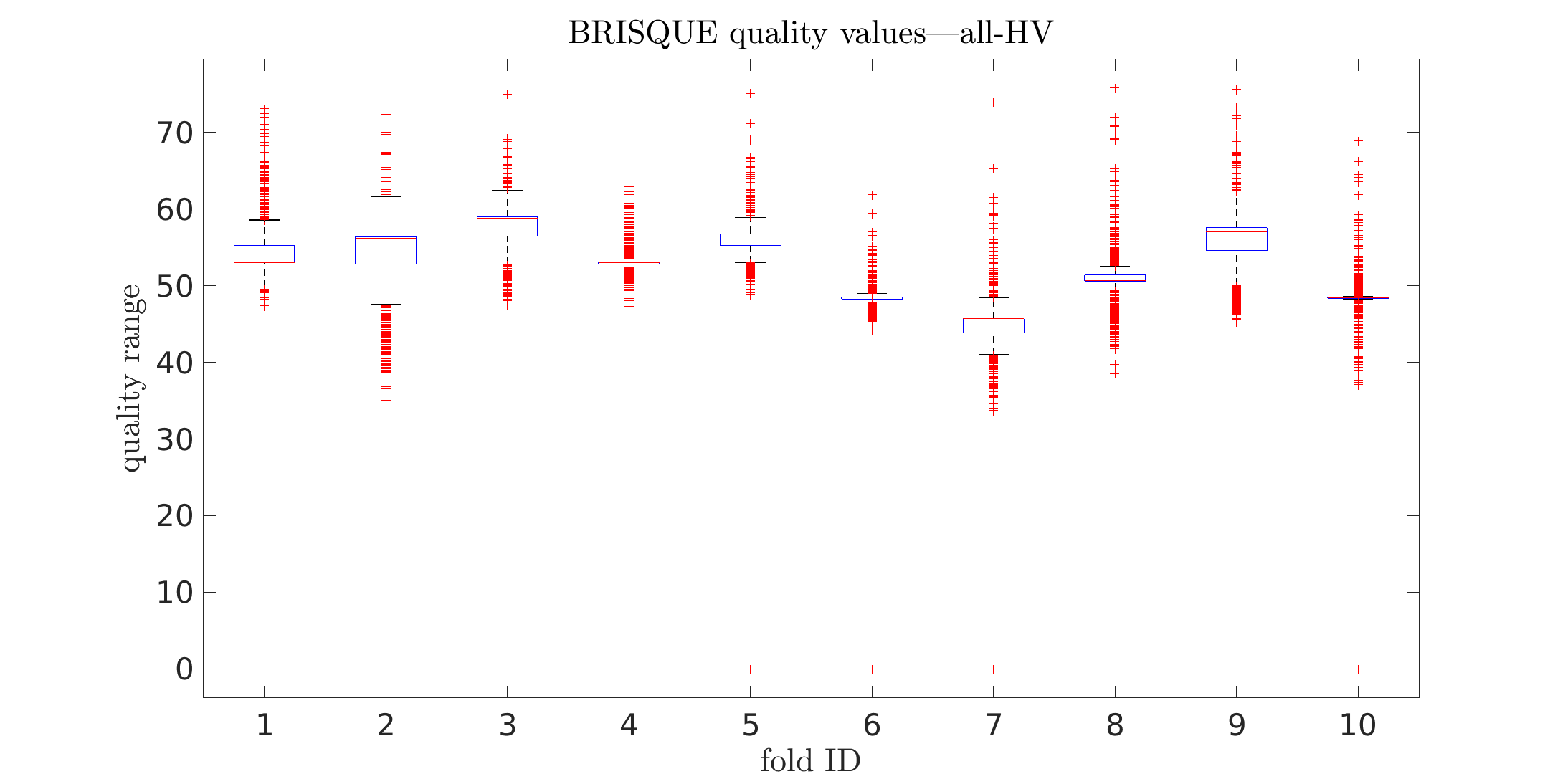}}
	\end{minipage} \vspace{0.5cm}  \\
	\begin{minipage}{0.33\textwidth}
	\centering{\includegraphics[width=1\textwidth]{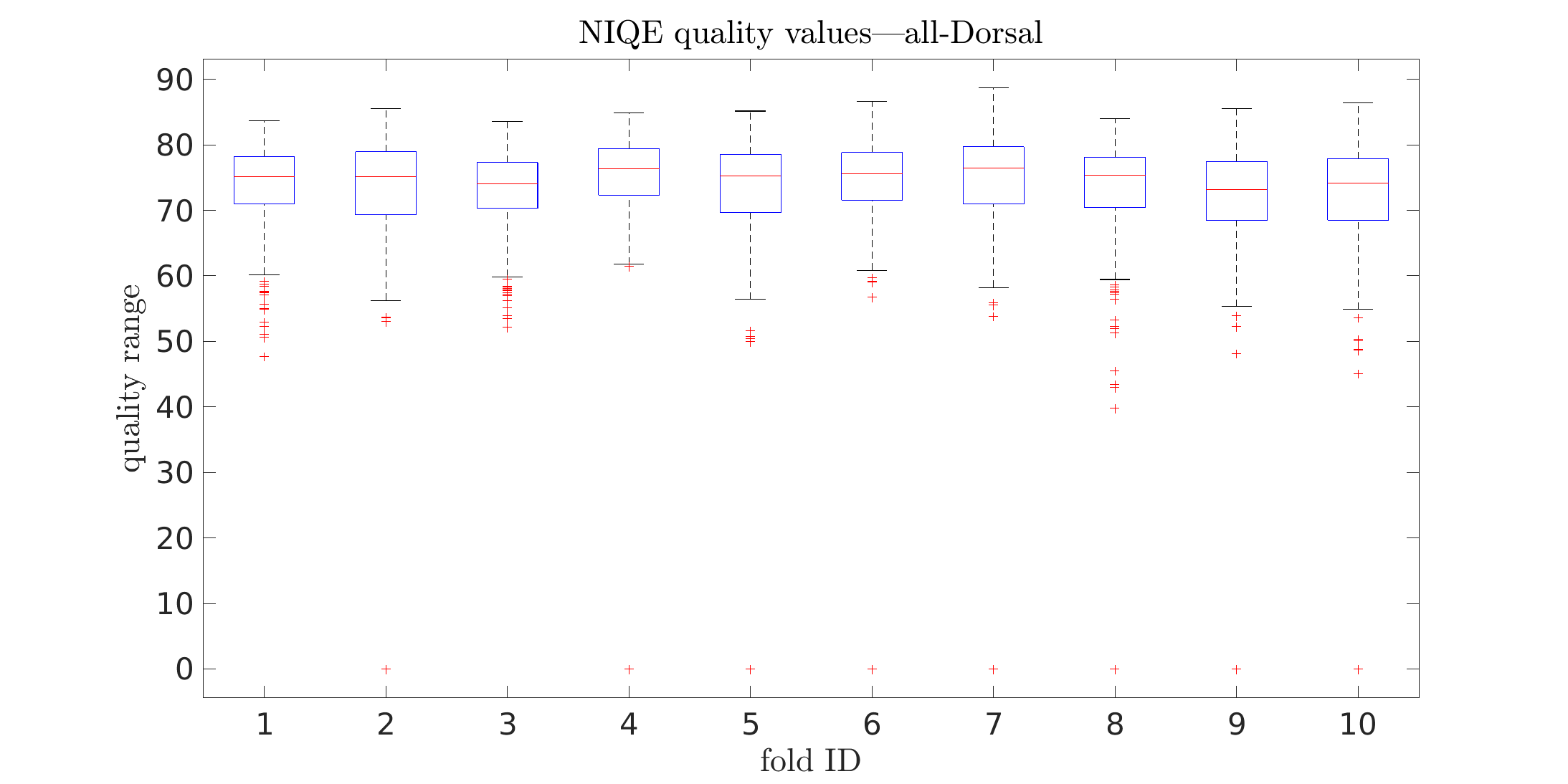}}
	\end{minipage}	
	\begin{minipage}{0.33\textwidth}
	\centering{\includegraphics[width=1\textwidth]{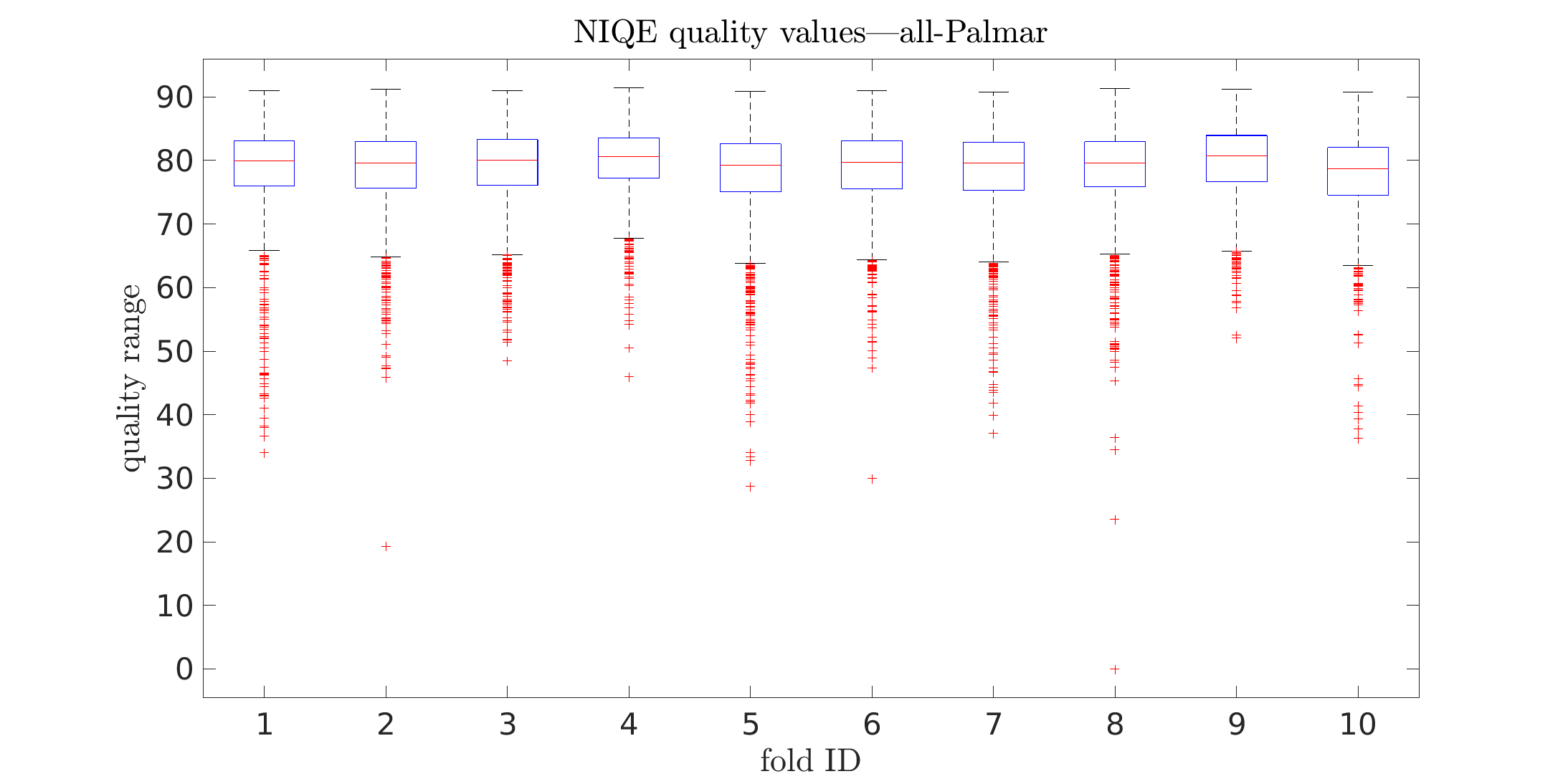}}
	\end{minipage} 
	\begin{minipage}{0.33\textwidth}
	\centering{\includegraphics[width=1\textwidth]{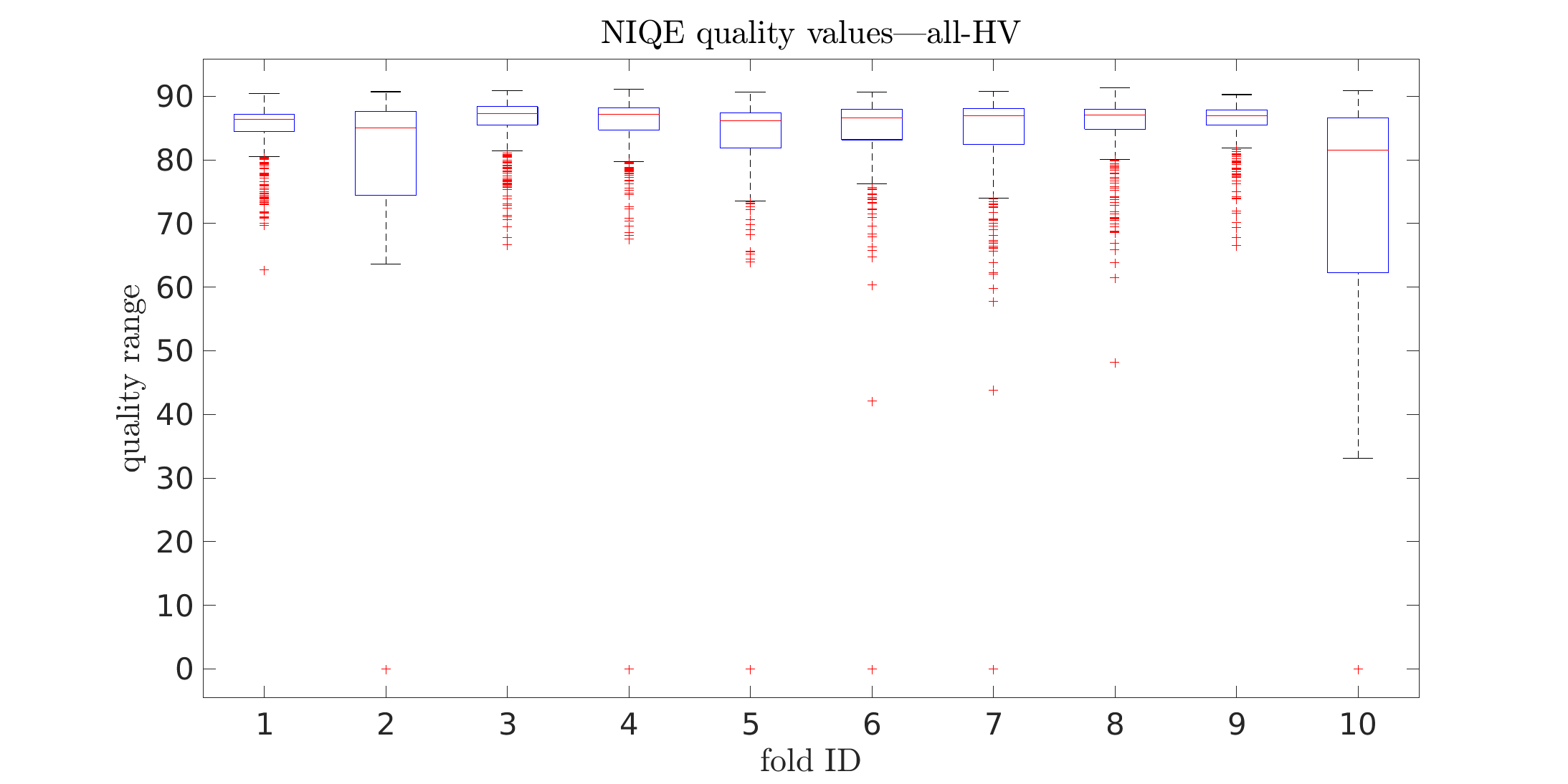}}
	\end{minipage}
	\caption{Quality results represented as boxplots using a 10 fold training's protocol for BRISQUE (first row) and NIQE (second row) utilising all data samples for dorsal (left column) or palmar (middle column) finger vein as well as hand vein (right column) databases.}
	\label{fig: training}
\end{figure*}

\begin{table}[ht!t] 
	\centering
	\caption{Overview on EER, FMR1000, ZeroFMR using finger vein datasets only.\label{tab: baselineFV}}
	{\begin{tabular*}{25pc}{@{\extracolsep{\fill}}ccccc@{}}\hline 
			database name & feature type & EER & FMR1000 & ZeroFMR\\
			\hline
			\textbf{finger vein - dorsal} & & & & \\ 
			\hline
			PLasDOR & GF & 0.0088 & 0.0166 & 0.0452 \\
			PLasDOR & MC & \textbf{0.0055} & \textbf{0.0083} & \textbf{0.0166} \\
			PLasDOR & SIFT & 0.0164 & 0.0508 & 0.2000 \\
			\hline
			PLEDDOR & GF & 0.0044 & 0.0080 & 0.0255 \\
			PLEDDOR & MC & \textbf{0.0013} & \textbf{0.0016} & \textbf{0.0061} \\
			PLEDDOR & SIFT & 0.0142 & 0.0669 & 0.4100 \\
			\hline
			\textbf{finger vein - palmar} & & & &\\ 
			\hline
			FV-USM&GF&0.0400&0.1250&0.2799 \\
			FV-USM&MC&\textbf{0.0261}&\textbf{0.0662}&\textbf{0.2516 }\\
			FV-USM&SIFT&0.0753&0.2442&0.4333 \\
			\hline
			HKPU-FV&GF&0.1124&0.3033&0.5163 \\
			HKPU-FV&MC&\textbf{0.1028}&\textbf{0.2387}&\textbf{0.3771} \\
			HKPU-FV&SIFT&0.1672&0.6417&0.9218 \\
			\hline
			MMCBNU&GF&0.1071&0.4126&0.7322 \\
			MMCBNU&MC&\textbf{0.0517}&\textbf{0.2042}&\textbf{0.5928} \\
			MMCBNU&SIFT&0.0809&0.9981&0.9997 \\
			\hline
			PLasPAL&GF&\textbf{0.0172}&\textbf{0.0241}&\textbf{0.0286} \\
			PLasPAL&MC&0.0213&0.0413&0.0663 \\
			PLasPAL&SIFT&0.0742&0.2788&0.6136 \\
			\hline
			PLEDPAL&GF&0.0038&0.0050&0.0097 \\
			PLEDPAL&MC&\textbf{0.0019}&\textbf{0.0019}&\textbf{0.0044} \\
			PLEDPAL&SIFT&0.0610&0.2786&0.9991 \\
			\hline
			SDUMLA&GF&0.1198&0.2803&0.6027 \\
			SDUMLA&MC&\textbf{0.0492}&\textbf{0.0975}&\textbf{0.5269} \\
			SDUMLA&SIFT&0.0884&0.2462&0.5007 \\
			\hline
			UTFVP&GF&0.0073&0.0203&0.0333 \\
			UTFVP&MC&\textbf{0.0069}&\textbf{0.0138}&\textbf{0.0337} \\
			UTFVP&SIFT&0.1190&0.5388&0.8157 \\
			\hline
	\end{tabular*}}{}
\end{table}

\begin{table}[ht!t] 
	\centering
	\caption{Overview on EER, FMR1000, ZeroFMR using hand vein datasets only.\label{tab: baselineHV}}
	{\begin{tabular*}{25pc}{@{\extracolsep{\fill}}ccccc@{}}\hline 
			database name & feature type & EER & FMR1000 & ZeroFMR\\
			\hline
			\textbf{hand vein} & & & &\\ 
			\hline 
			CIE&GF&0.0532&0.0842&0.1696 \\
			CIE&MC&\textbf{0.0433}&\textbf{0.0535}&\textbf{0.0600} \\
			CIE&SIFT&0.2642&0.7642&0.8471 \\
			\hline
			PRefl&GF&0.0645&0.4601&0.8276 \\
			PRefl&MC&0.0348&0.1358&0.2079 \\
			PRefl&SIFT&\textbf{0.0125}&\textbf{0.0178}&\textbf{0.0212} \\
			\hline
			PTrans&GF&0.0500&0.3125&0.7975 \\
			PTrans&MC&0.0292&0.1270&0.6697 \\
			PTrans&SIFT&\textbf{0.0210}&\textbf{0.0461}&\textbf{0.5026} \\
			\hline
			VERA&GF&\textbf{0.0193}&\textbf{0.0413}&\textbf{0.0477} \\
			VERA&MC&0.0310&0.0446&0.0506 \\
			VERA&SIFT&0.0340&0.1126&0.1911 \\
			\hline
	\end{tabular*}}{}
\end{table}

\section{Experimental Evaluation}\label{sec: evaluation}
In the following subsections, the results of the baseline performance evaluation of the selected databases (intending to establish a comparison benchmark for subsequently conducted experiments), the results of the extended evaluation based on \cite{Remy22b} and the findings of the proposed DL quality assessment method are presented and discussed.  

\begin{figure*}[!b]
	\begin{minipage}{0.33\textwidth}
	\centering{\includegraphics[width=1\textwidth]{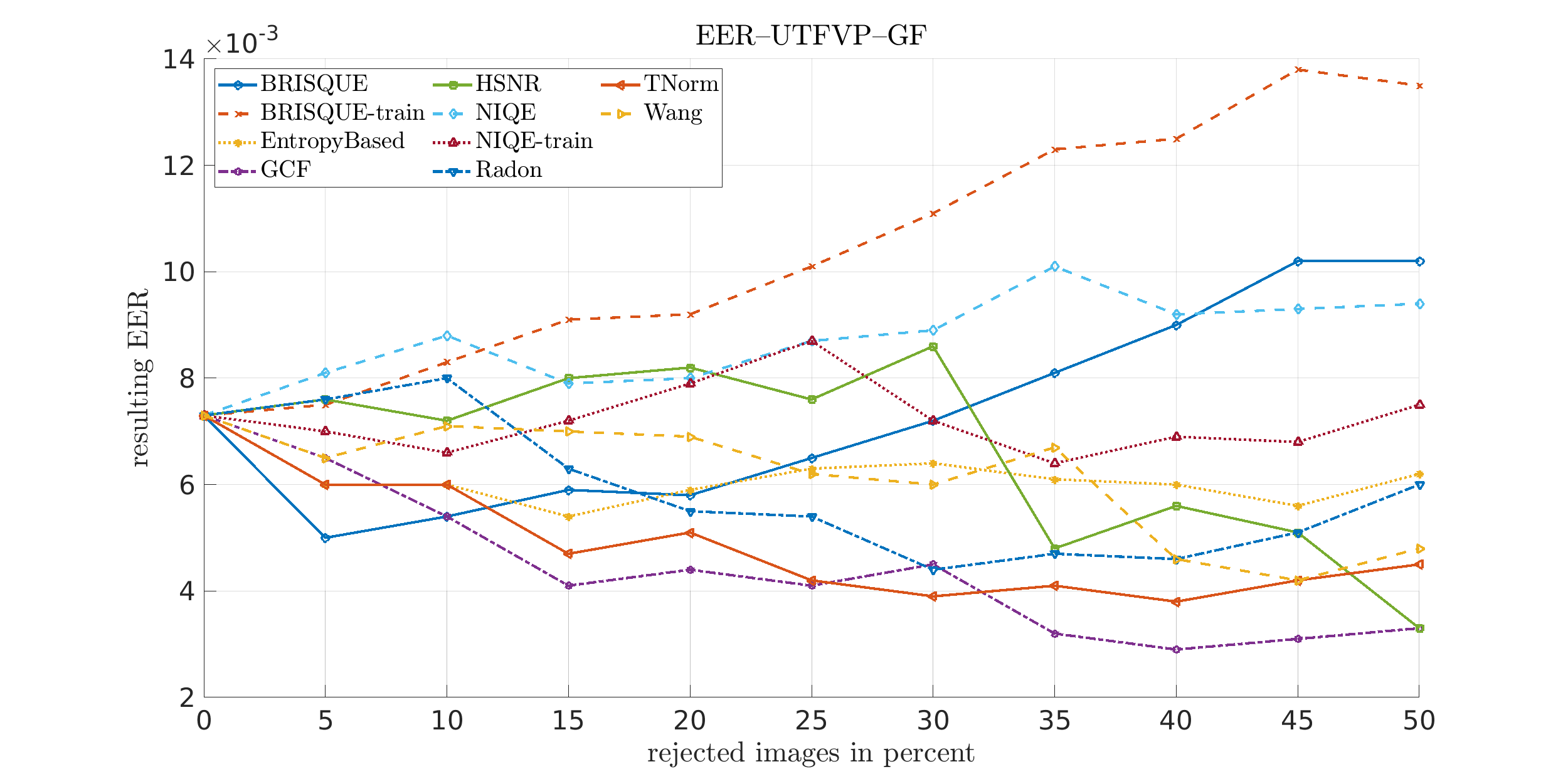}}
	\end{minipage}
	\begin{minipage}{0.33\textwidth}
	\centering{\includegraphics[width=1\textwidth]{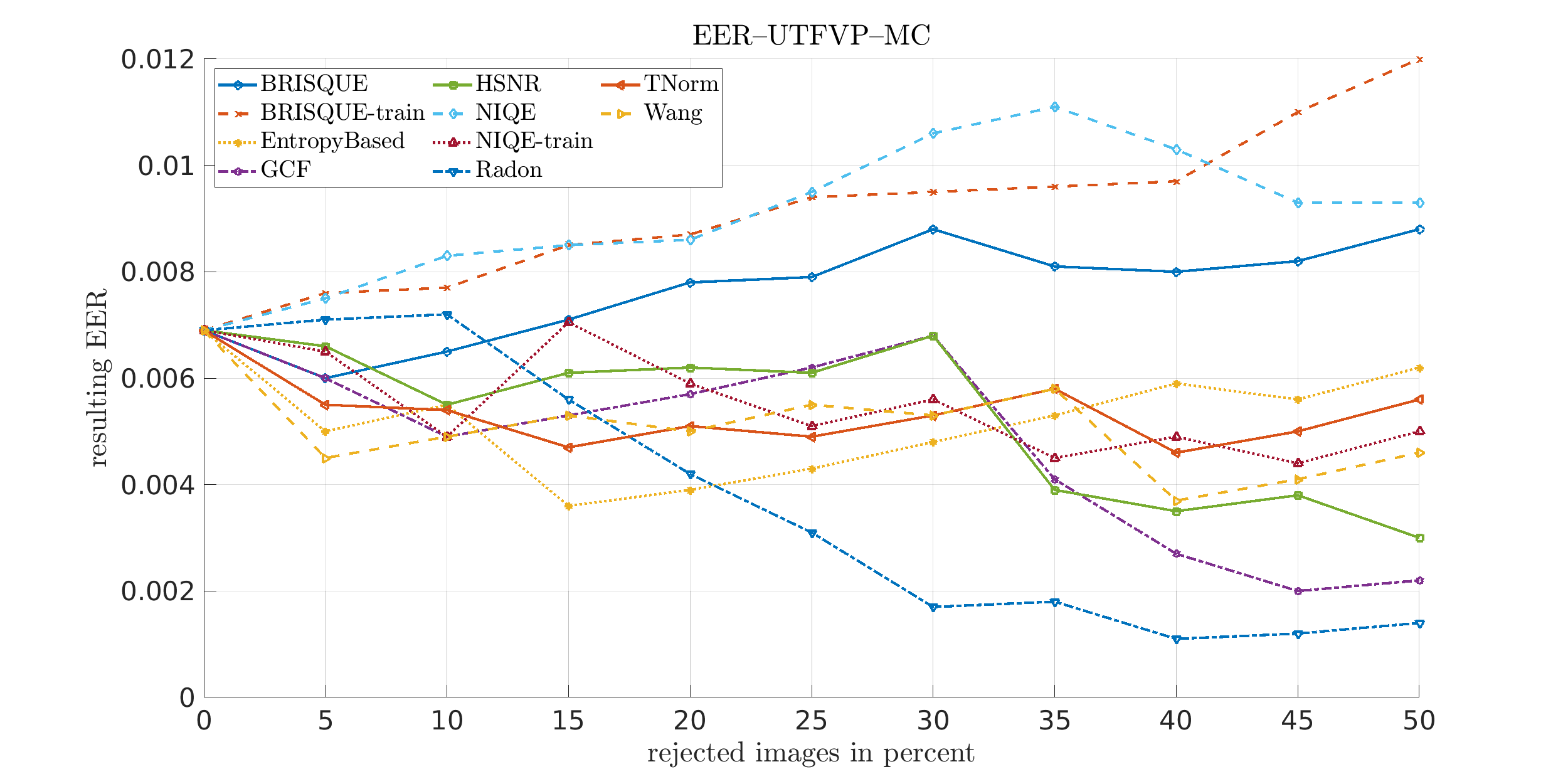}}
	\end{minipage}
	\begin{minipage}{0.33\textwidth}
	\centering{\includegraphics[width=1\textwidth]{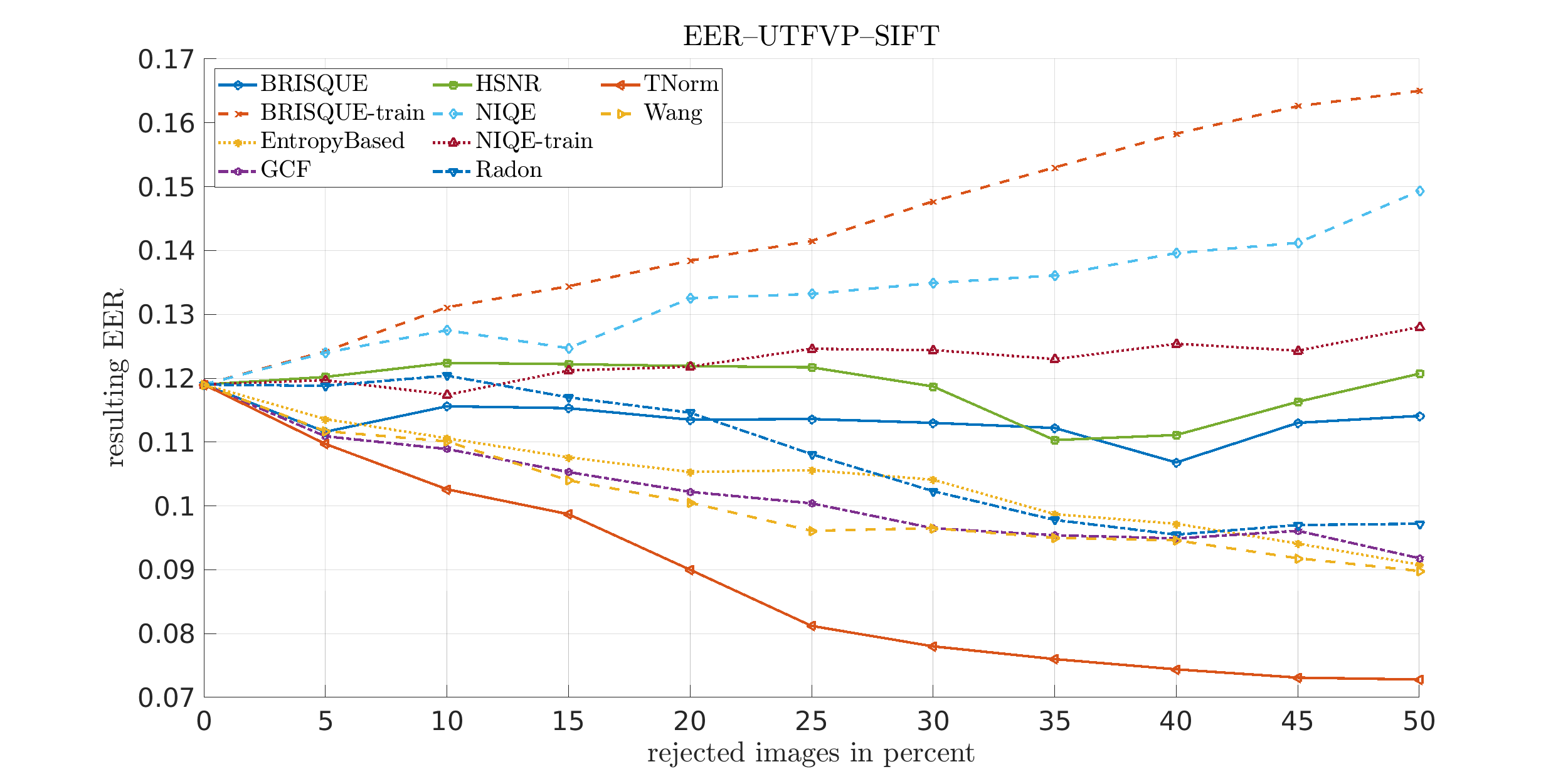}}
	\end{minipage} \vspace{0.5cm} \\
	\begin{minipage}{0.33\textwidth}
	\centering{\includegraphics[width=1\textwidth]{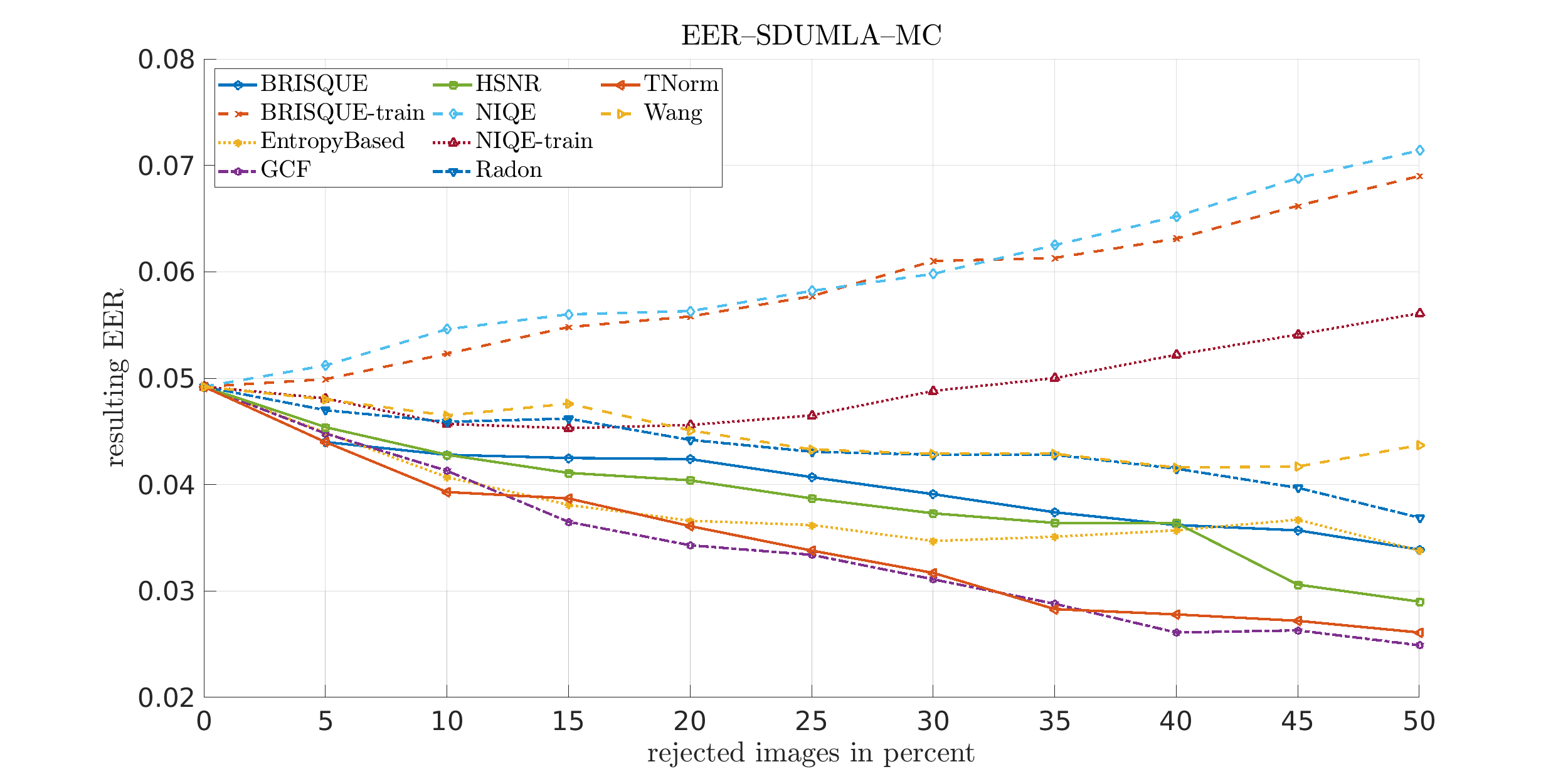}}
	\end{minipage} 
	\begin{minipage}{0.33\textwidth}
	\centering{\includegraphics[width=1\textwidth]{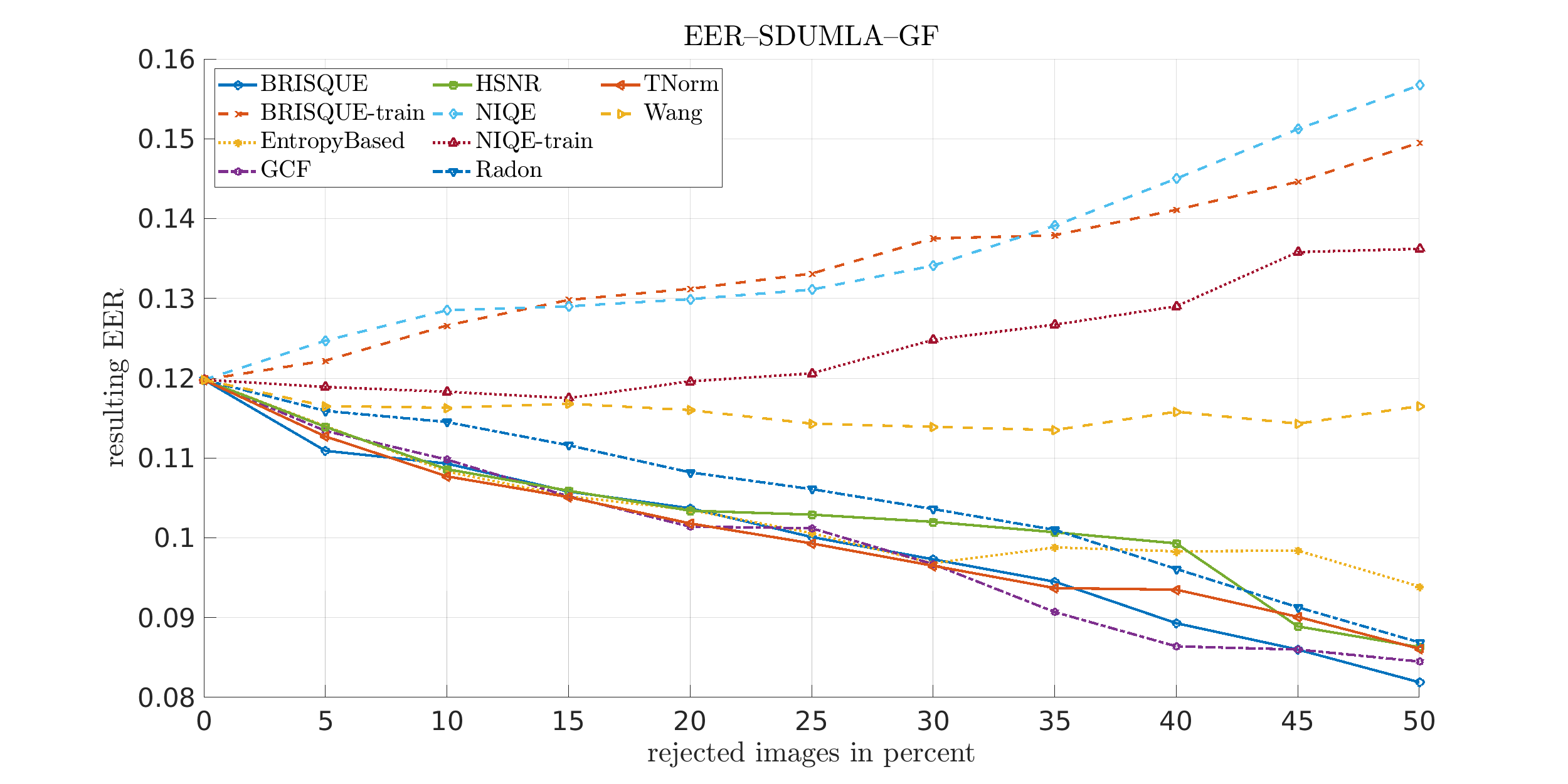}}
	\end{minipage} 
	\begin{minipage}{0.33\textwidth}
	\centering{\includegraphics[width=1\textwidth]{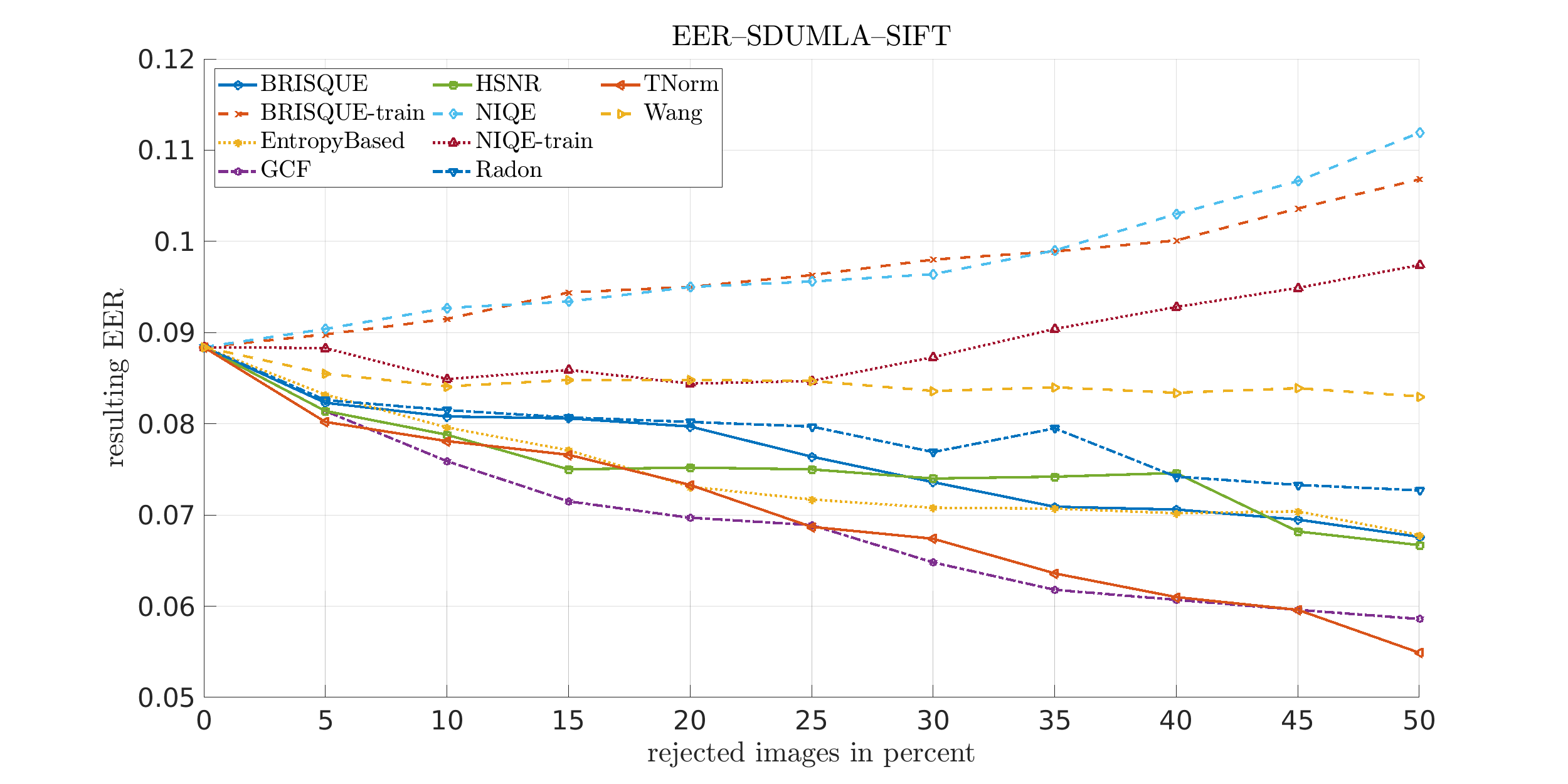}}
	\end{minipage} \\
	\caption{EER trend with increasing rate of rejected images on UTFVP (first row) and SDUMLA (second row).}
	\label{fig: BIOSIGvgl}
\end{figure*}

\subsection{Baseline Performance Evaluation}
Tables \ref{tab: baselineFV} and \ref{tab: baselineHV} list the corresponding recognition performance results in terms of the EER, the FMR1000 and the ZeroFMR for the finger as well as the hand databases, respectively, obtained by using FVC verification mode of the PLUS OpenVein Finger- and Hand-Vein Toolkit \cite{Kauba19b}. In both tables the best performance values for each database are highlighted in bold. These results serve as baseline to enable a subsequent analysis of the recognition performance progress. If a portion of the lowest quality images is successively removed from the evaluation database the performance should improve over the baseline one. \\
Tables \ref{tab: baselineFV} and \ref{tab: baselineHV} show, that MC as feature extraction achieves the best recognition performance, especially for the finger vein databases. The only exception is the PLasPAL where the application of GF was superior than MC and SIFT. For the hand vein databases in two of four cases SIFT was the best feature extraction method, while MC and GF choice was best in one case each. \\
The overall recognition performance is on a very high level, with an EER of close to zero in several cases for most databases. Thus, it is assumed that the subsequently performed quality based experiments, where the lowest quality images are successively discarded from the performance evaluation, will not significantly improve the EER values, but might result in an improvement of the FMR1000 and ZeroFMR. 

\subsection{Quality evaluation using extended training data}
In this extension of \cite{Remy22b} BRISQUE and NIQE were re-trained using all databases mentioned in Section \ref{sec: datasets}. The evaluation results of the 10-fold BRISQUE and NIQE training are depicted in Figure \ref{fig: training}, showing the whole range of the quality values for all dorsal (left column), palmar (middle column) and hand vein (right column) samples, respectively, across all three quality classes. This statistical evaluation was only performed for BRISQUE and NIQE to highlight their stability or possible variations regardless which fold is used. Ideally, the quality values should remain stable across all folds. It can be seen that the training of BRISQUE (first row) and NIQE (second row) highly corresponds to the databases used. While, for NIQE the boxplots representing the obtained quality values are quite stable (independent of the selected fold), BRISQUE exhibits more variation, especially using dorsal finger vein data samples. Hence, NIQE should also achieve more stable results as compared to BRISQUE for the leave one database out experiments. Furthermore, the overall quality values for NIQE are expected to be higher than for BRISQUE, which is a general trend as NIQE is trained on high quality images only.

\begin{figure*}[!t]
	\begin{minipage}{0.33\textwidth}
	\centering{\includegraphics[width=1\textwidth]{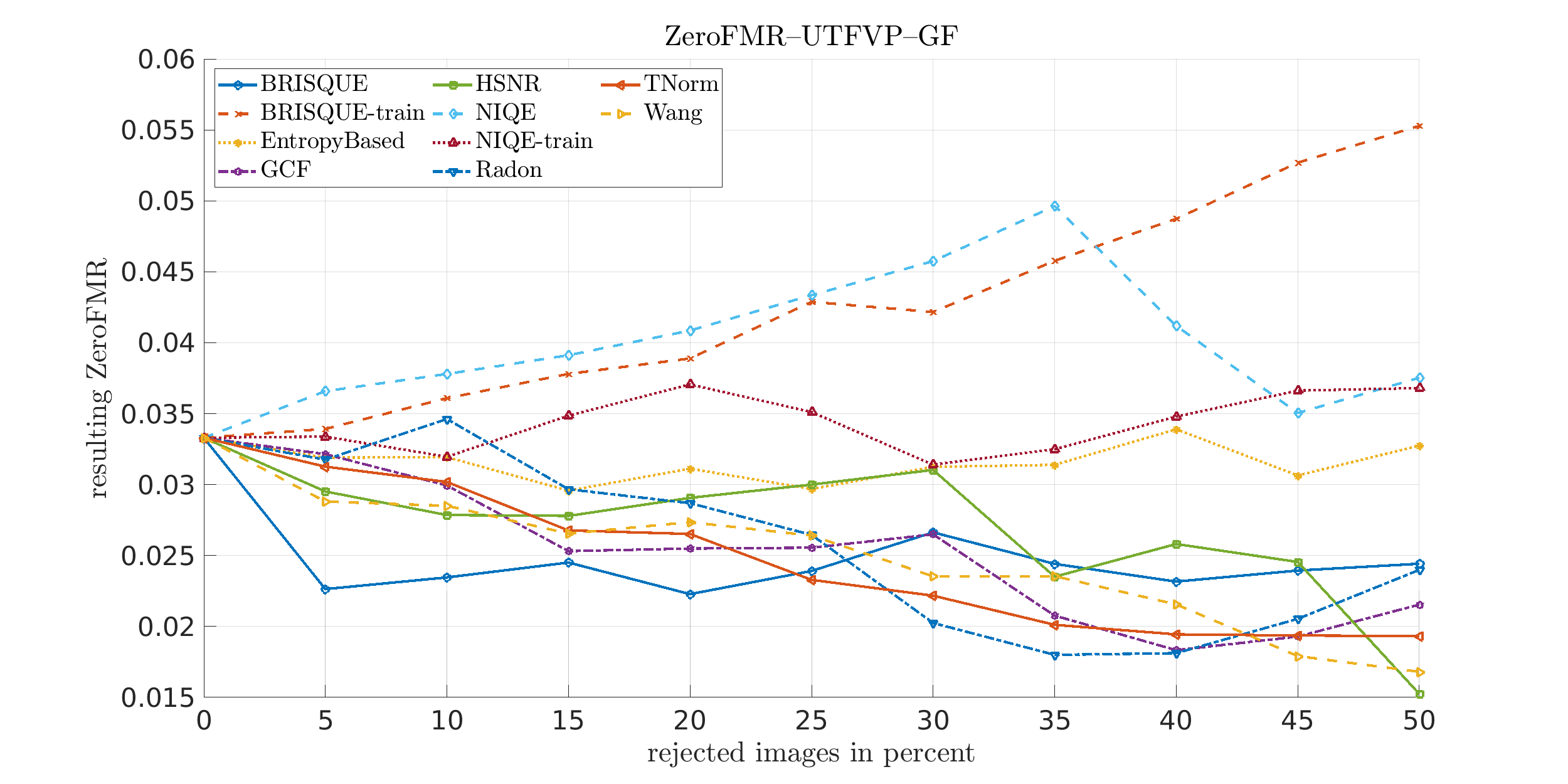}}
	\end{minipage}
	\begin{minipage}{0.33\textwidth}
	\centering{\includegraphics[width=1\textwidth]{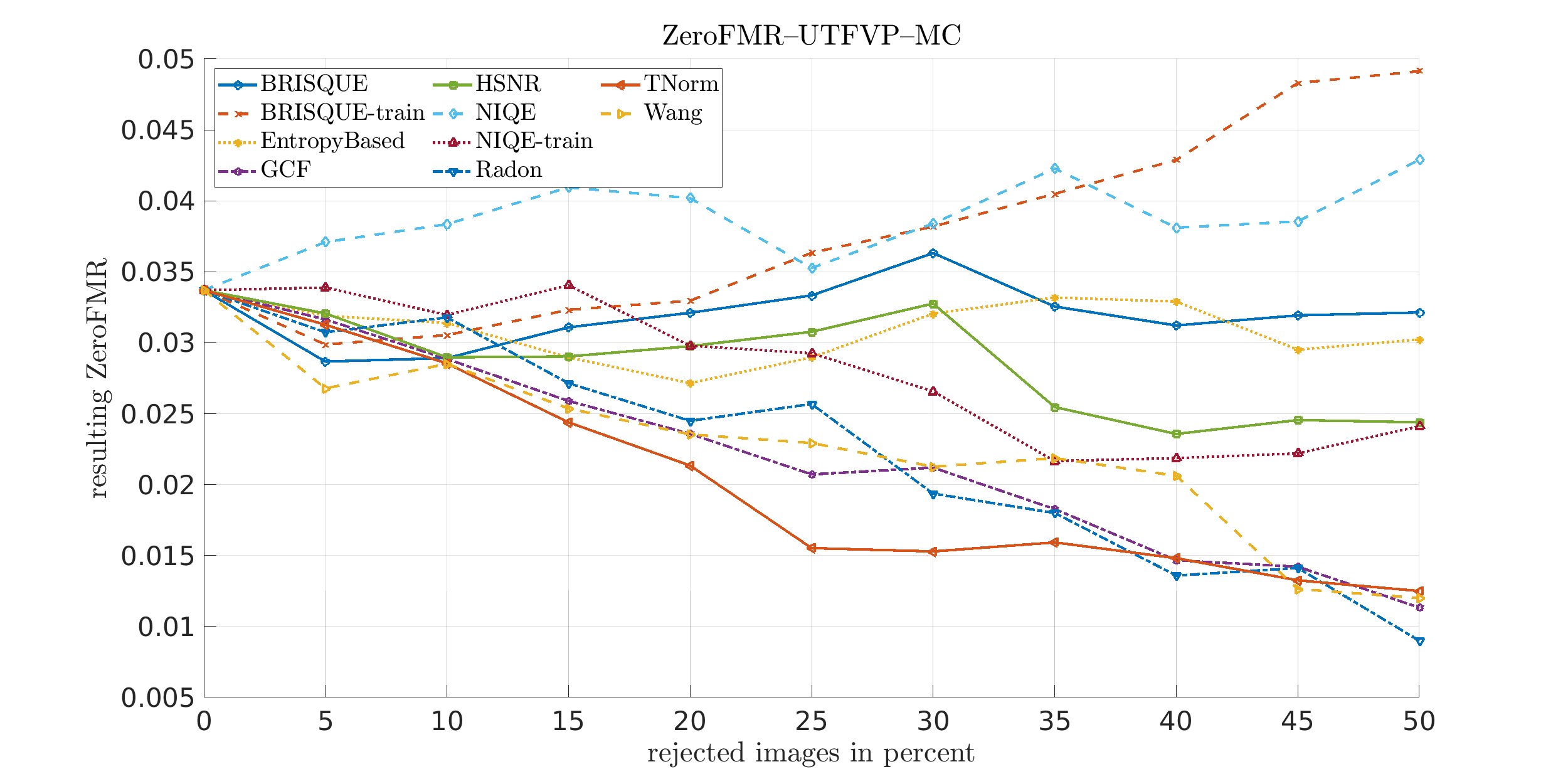}}
	\end{minipage}
	\begin{minipage}{0.33\textwidth}
	\centering{\includegraphics[width=1\textwidth]{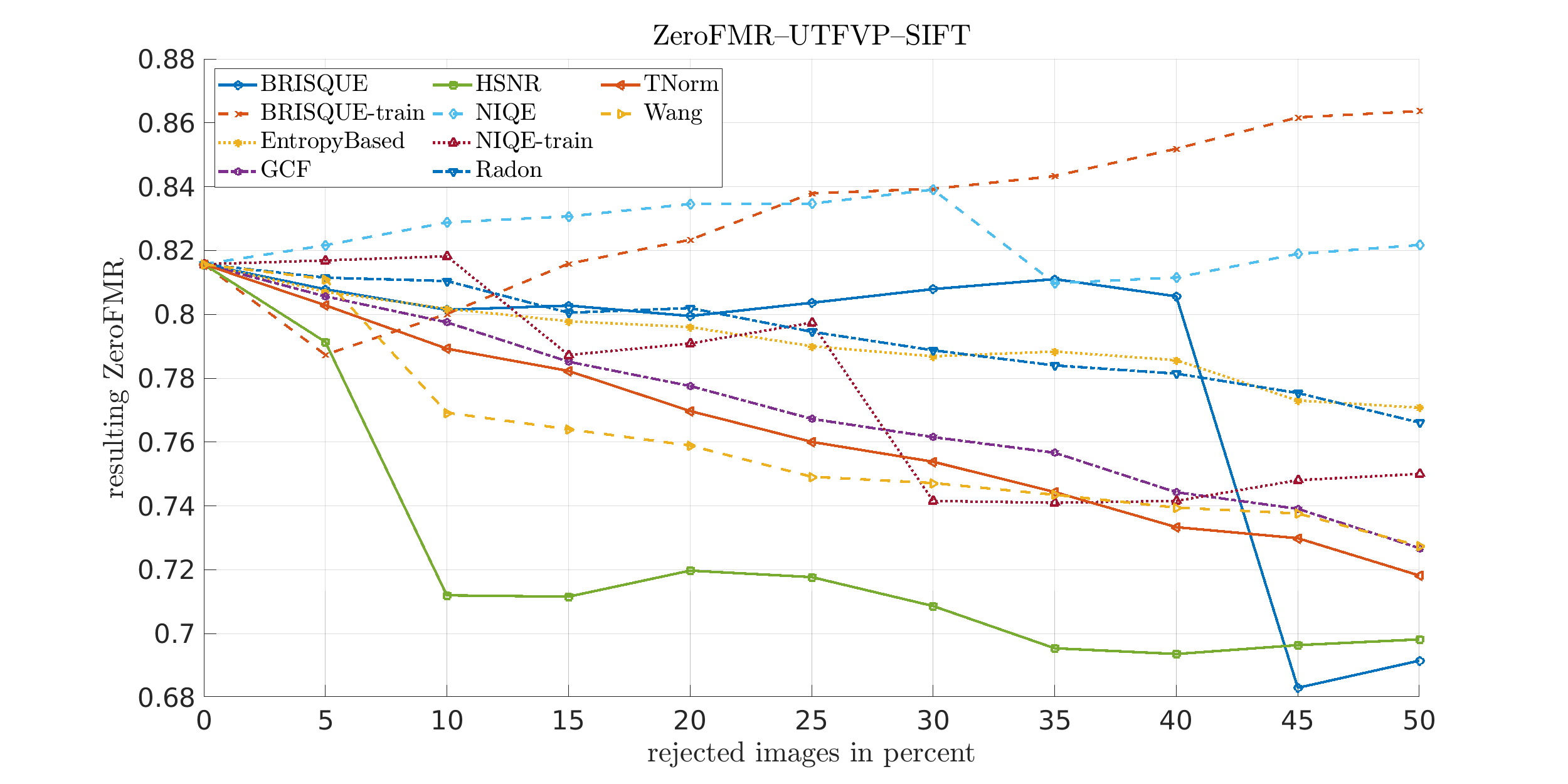}}
	\end{minipage} \vspace{0.5cm} \\
	\begin{minipage}{0.33\textwidth}
	\centering{\includegraphics[width=1\textwidth]{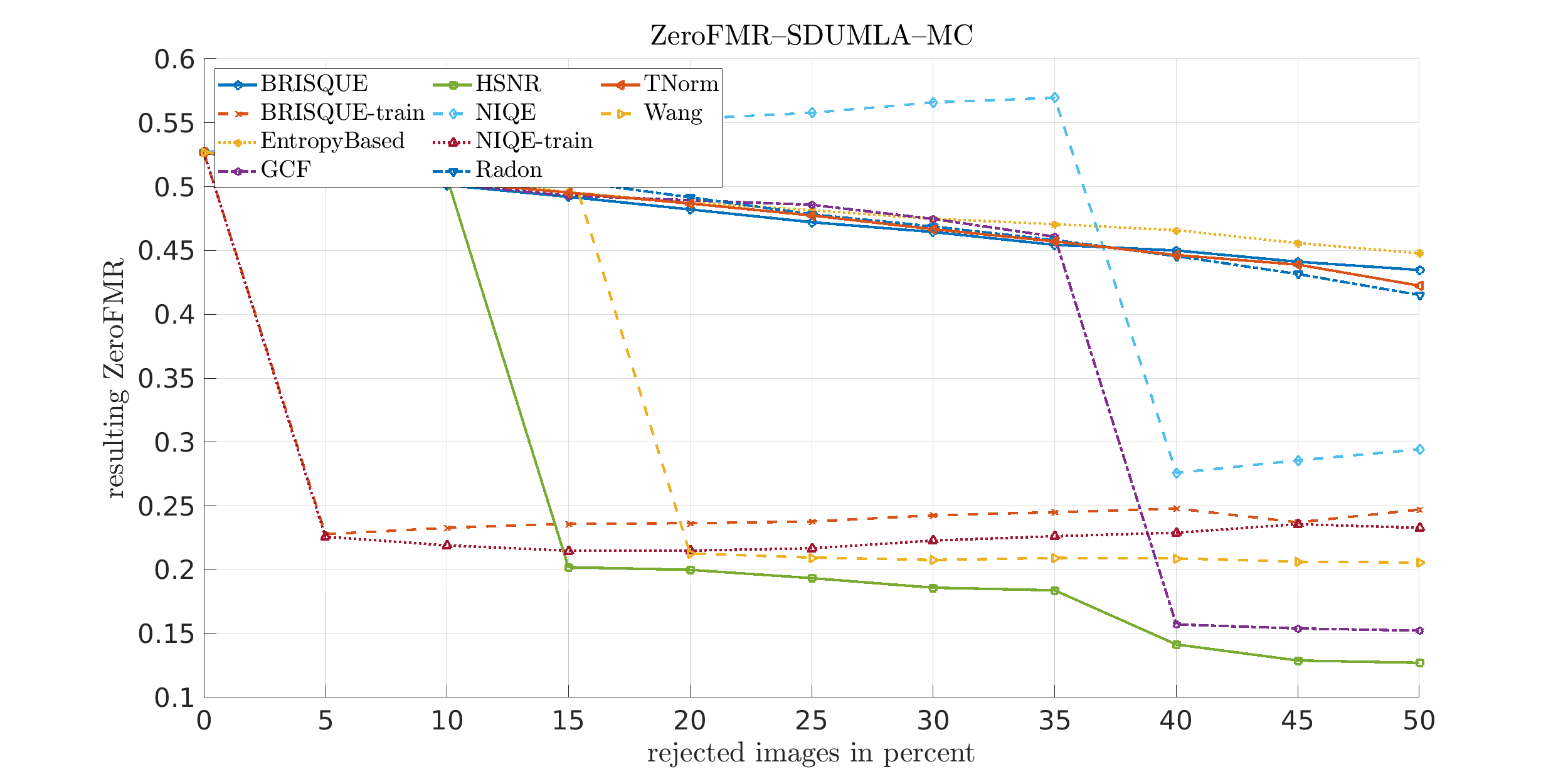}}
	\end{minipage} 
	\begin{minipage}{0.33\textwidth}
	\centering{\includegraphics[width=1\textwidth]{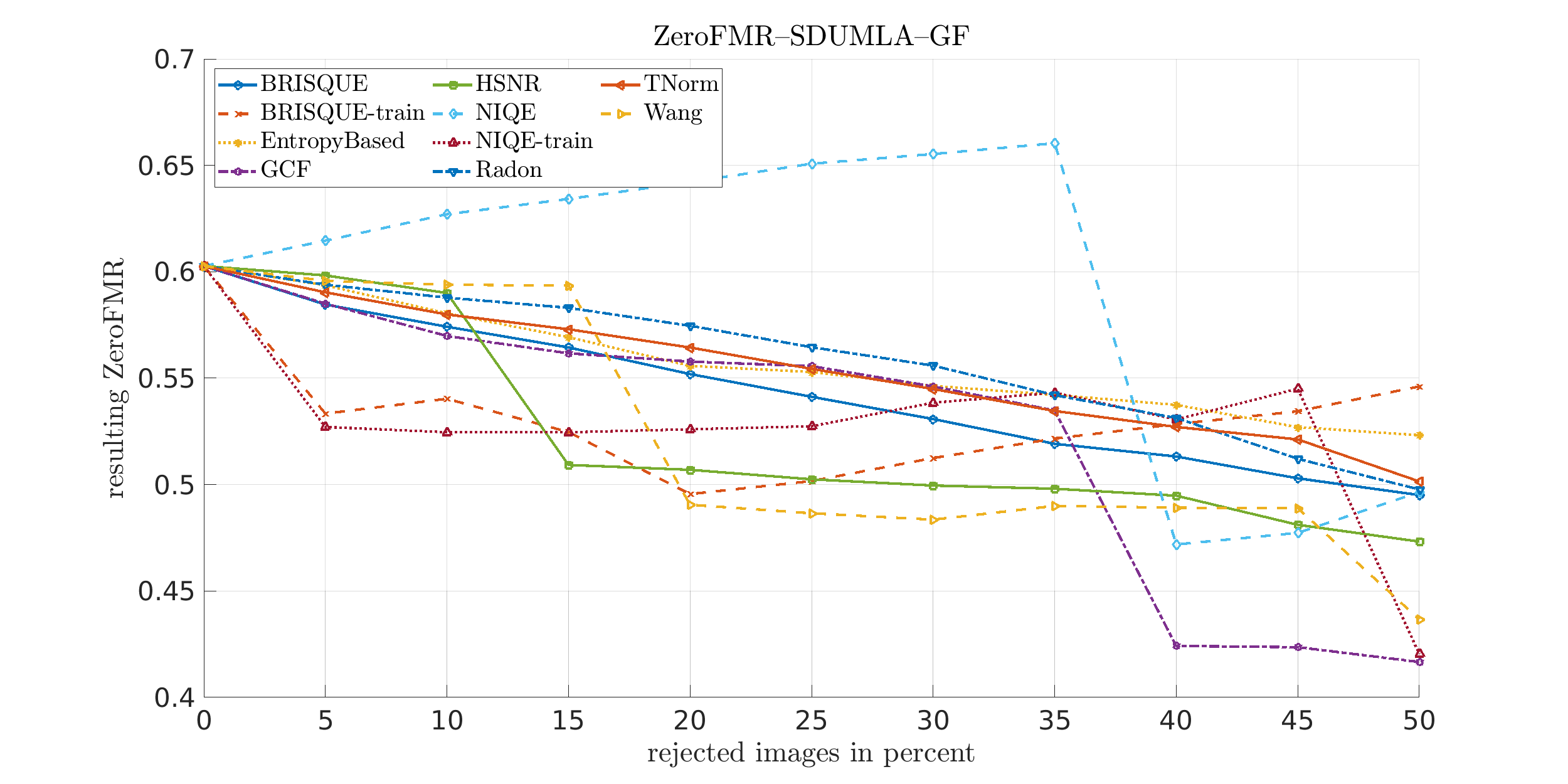}}
	\end{minipage} 
	\begin{minipage}{0.33\textwidth}
	\centering{\includegraphics[width=1\textwidth]{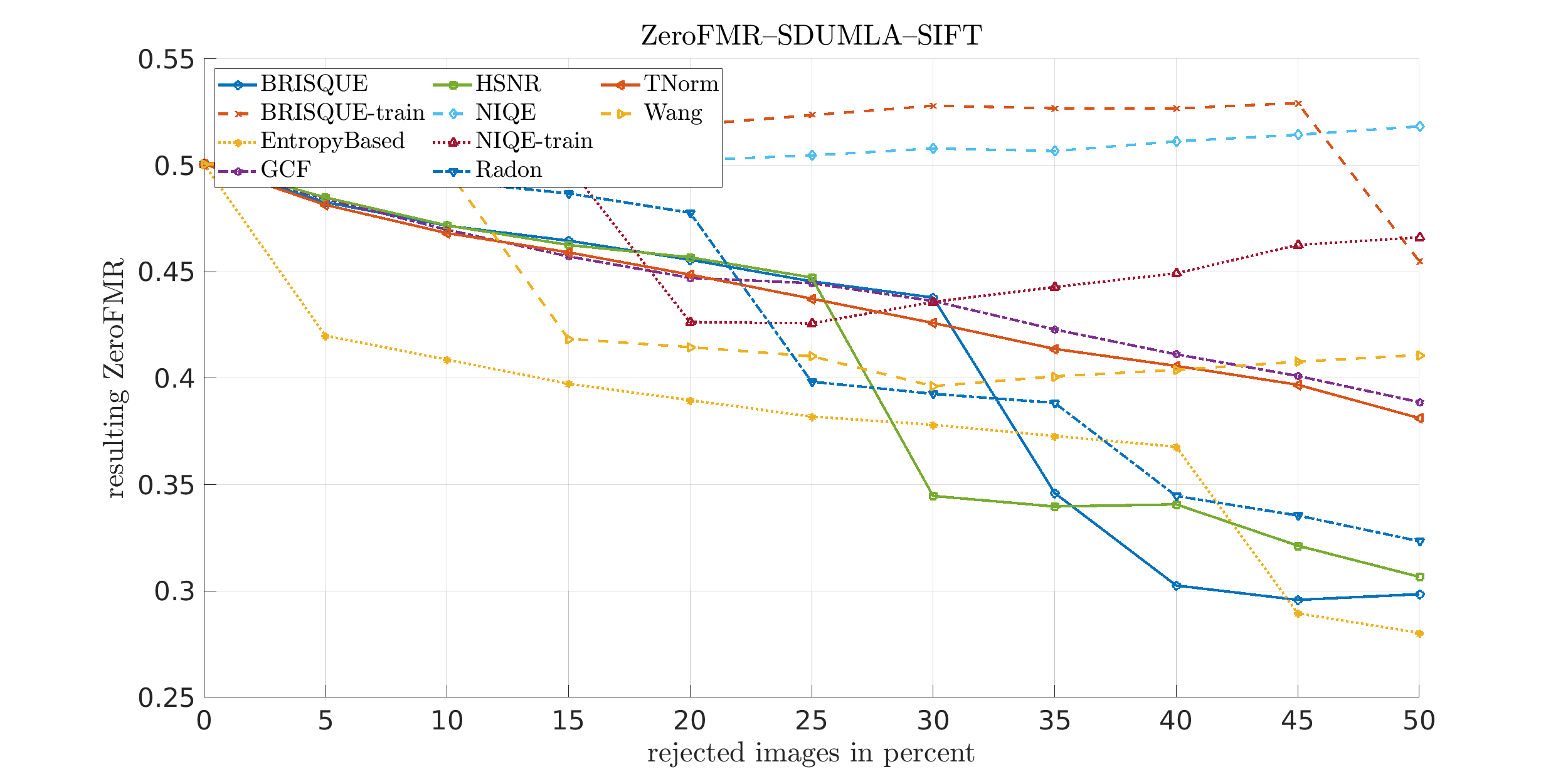}}
	\end{minipage} \\
	\caption{ZeroFMR trend with increasing rate of rejected images on UTFVP (first row) and SDUMLA (second row).}
	\label{fig: BIOSIGvgl_Zero}
\end{figure*}

\begin{figure*}[!t]
	\begin{minipage}{0.33\textwidth}
	\centering{\includegraphics[width=1\textwidth]{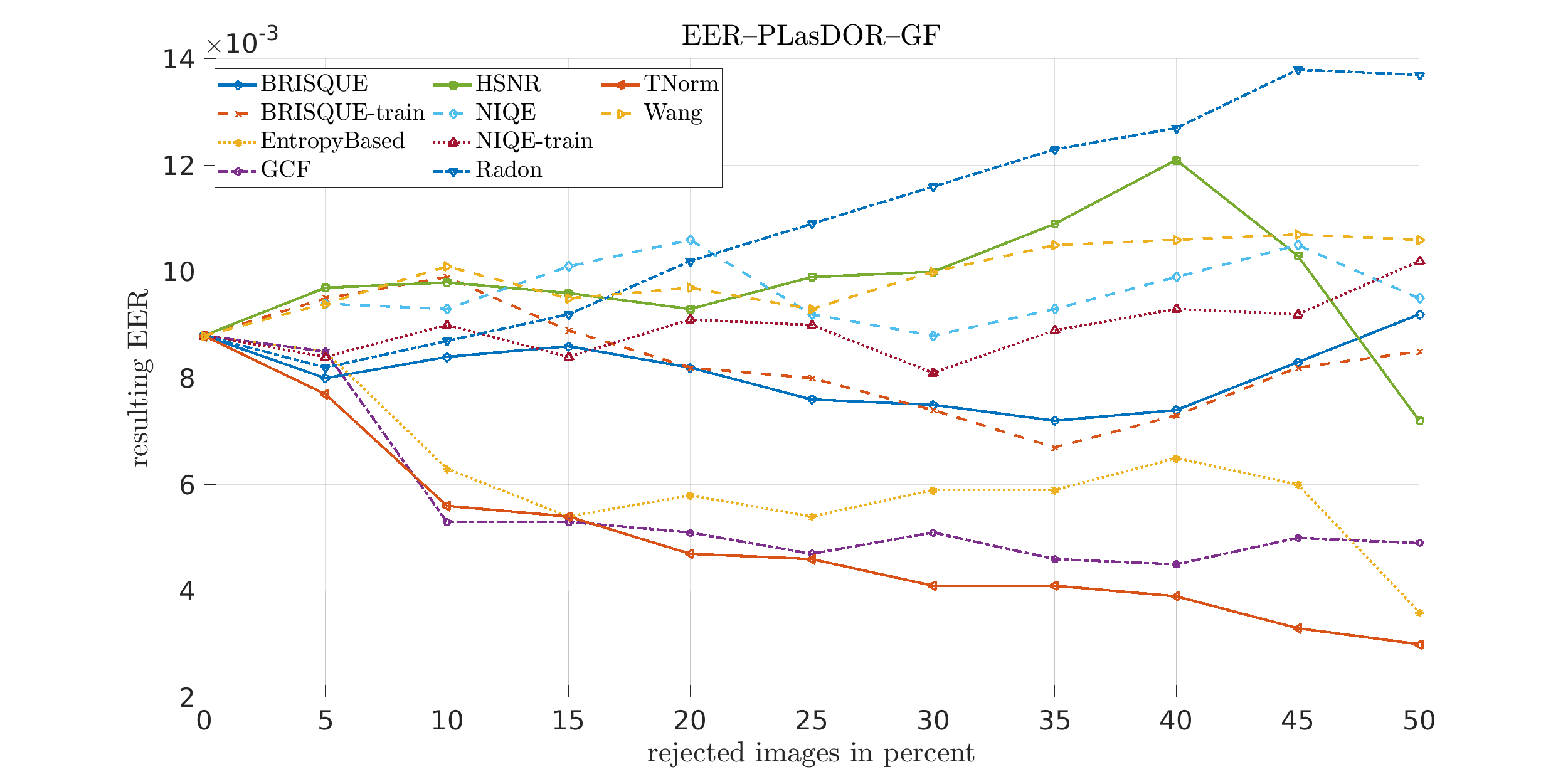}}
	\end{minipage}
	\begin{minipage}{0.33\textwidth}
	\centering{\includegraphics[width=1\textwidth]{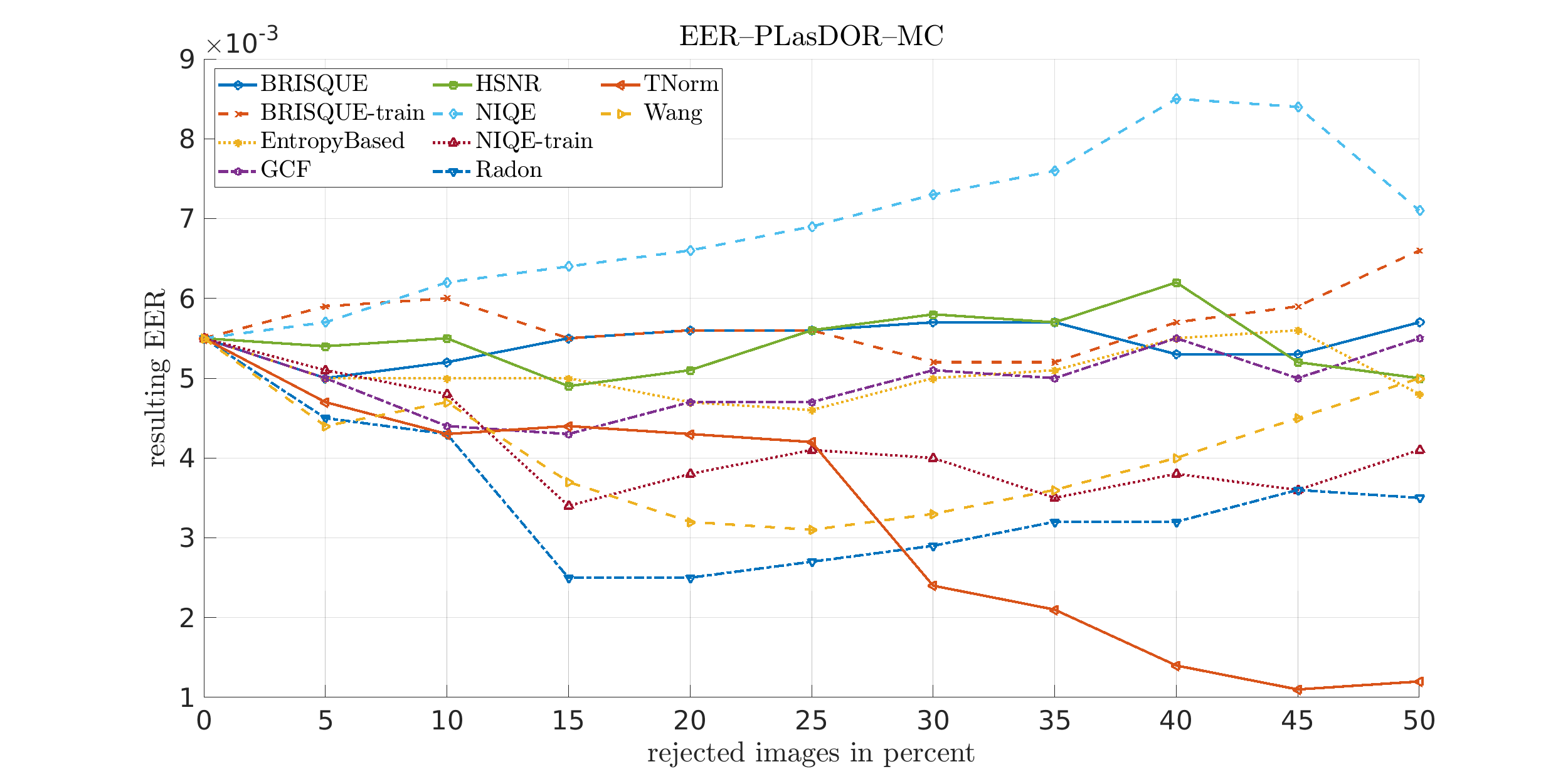}}
	\end{minipage}
	\begin{minipage}{0.33\textwidth}
	\centering{\includegraphics[width=1\textwidth]{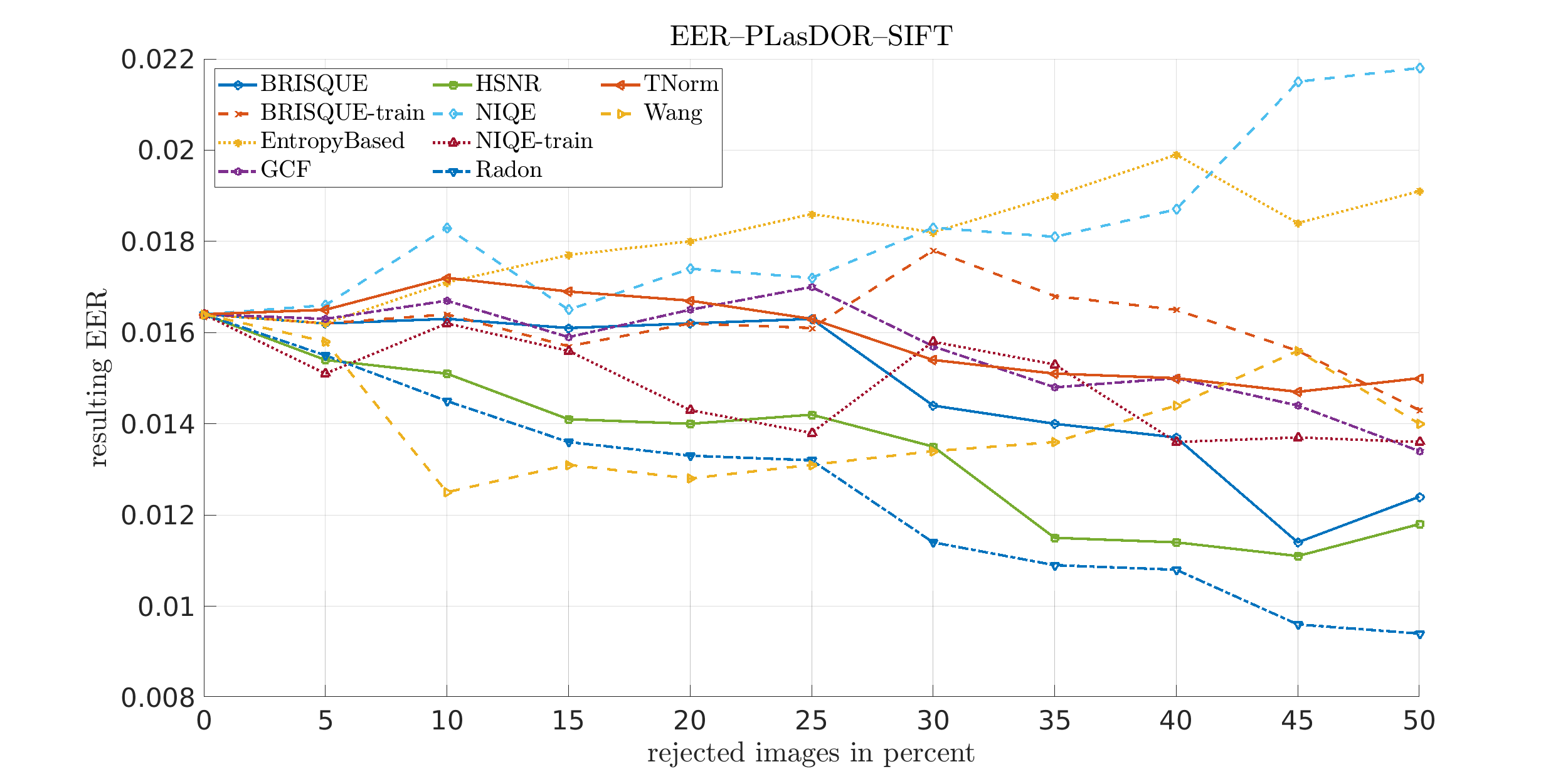}}
	\end{minipage} \vspace{0.5cm} \\
	\begin{minipage}{0.33\textwidth}
	\centering{\includegraphics[width=1\textwidth]{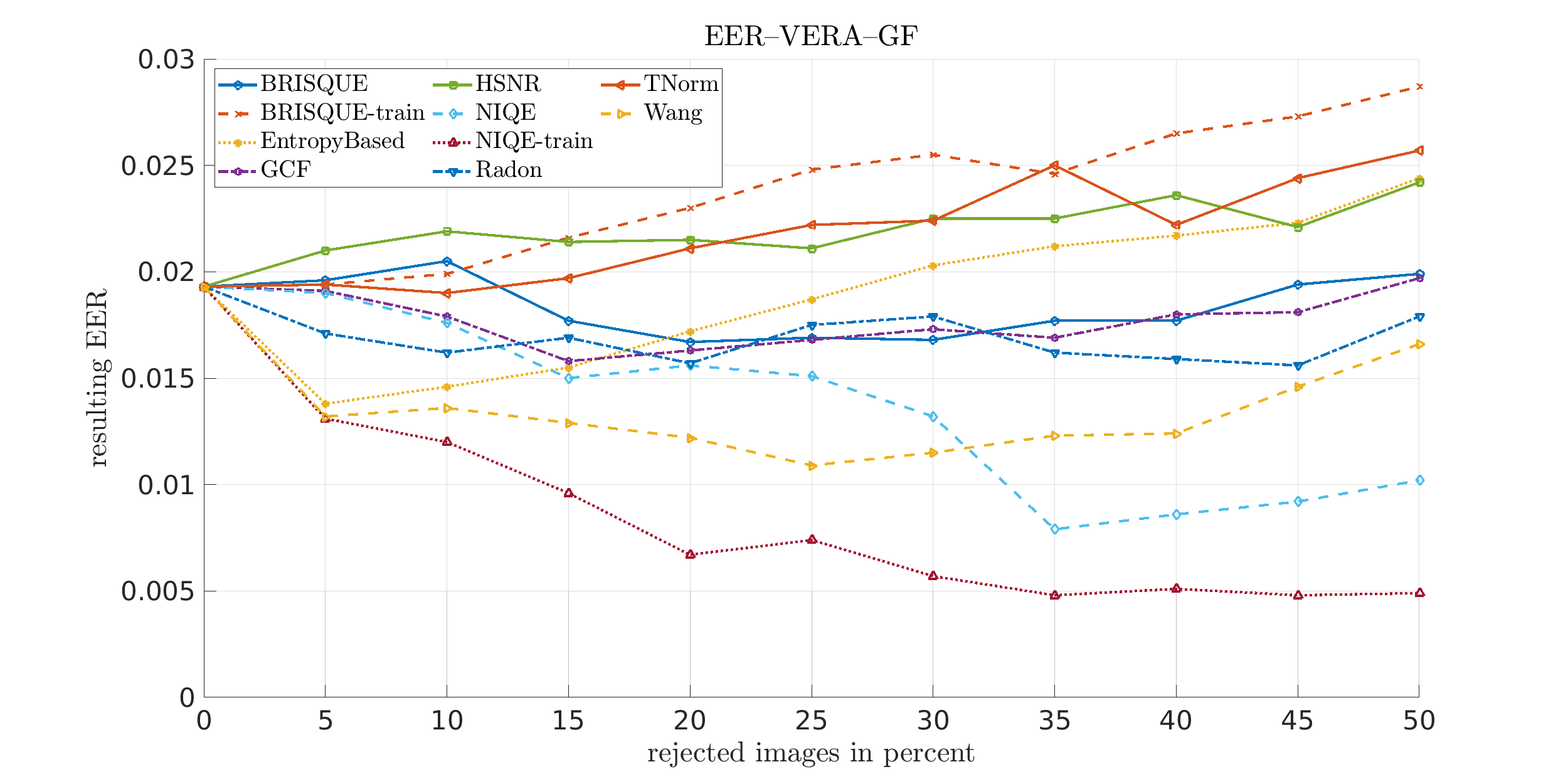}}
	\end{minipage} 
	\begin{minipage}{0.33\textwidth}
	\centering{\includegraphics[width=1\textwidth]{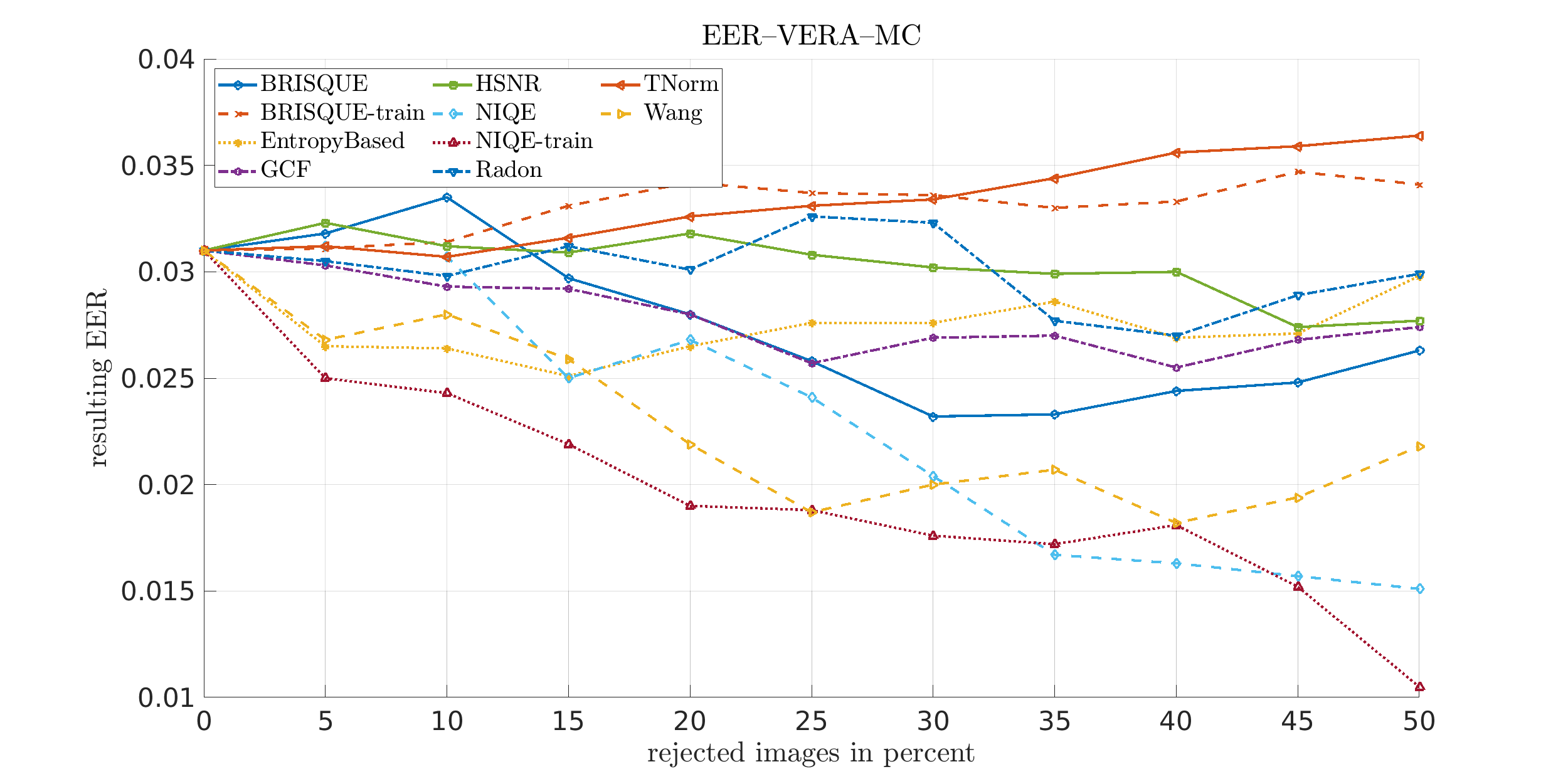}}
	\end{minipage} 
	\begin{minipage}{0.33\textwidth}
	\centering{\includegraphics[width=1\textwidth]{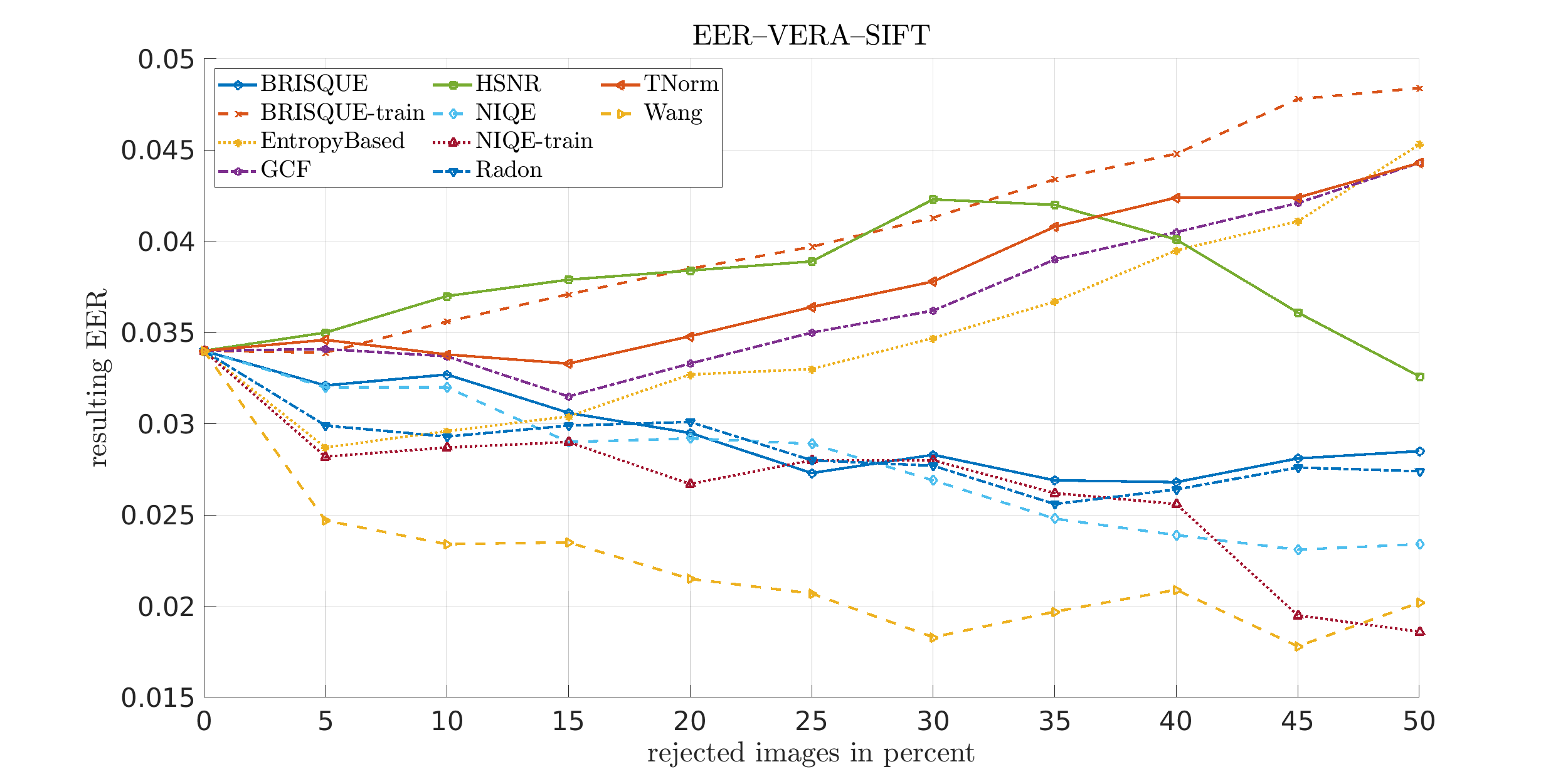}}
	\end{minipage} \\
	\caption{EER trend with increasing rate of rejected images on PLasDOR (first row) and VERA (second row).}
	\label{fig: NEWvgl}
\end{figure*}

\begin{figure*}[!t]
	\begin{minipage}{0.33\textwidth}
	\centering{\includegraphics[width=1\textwidth]{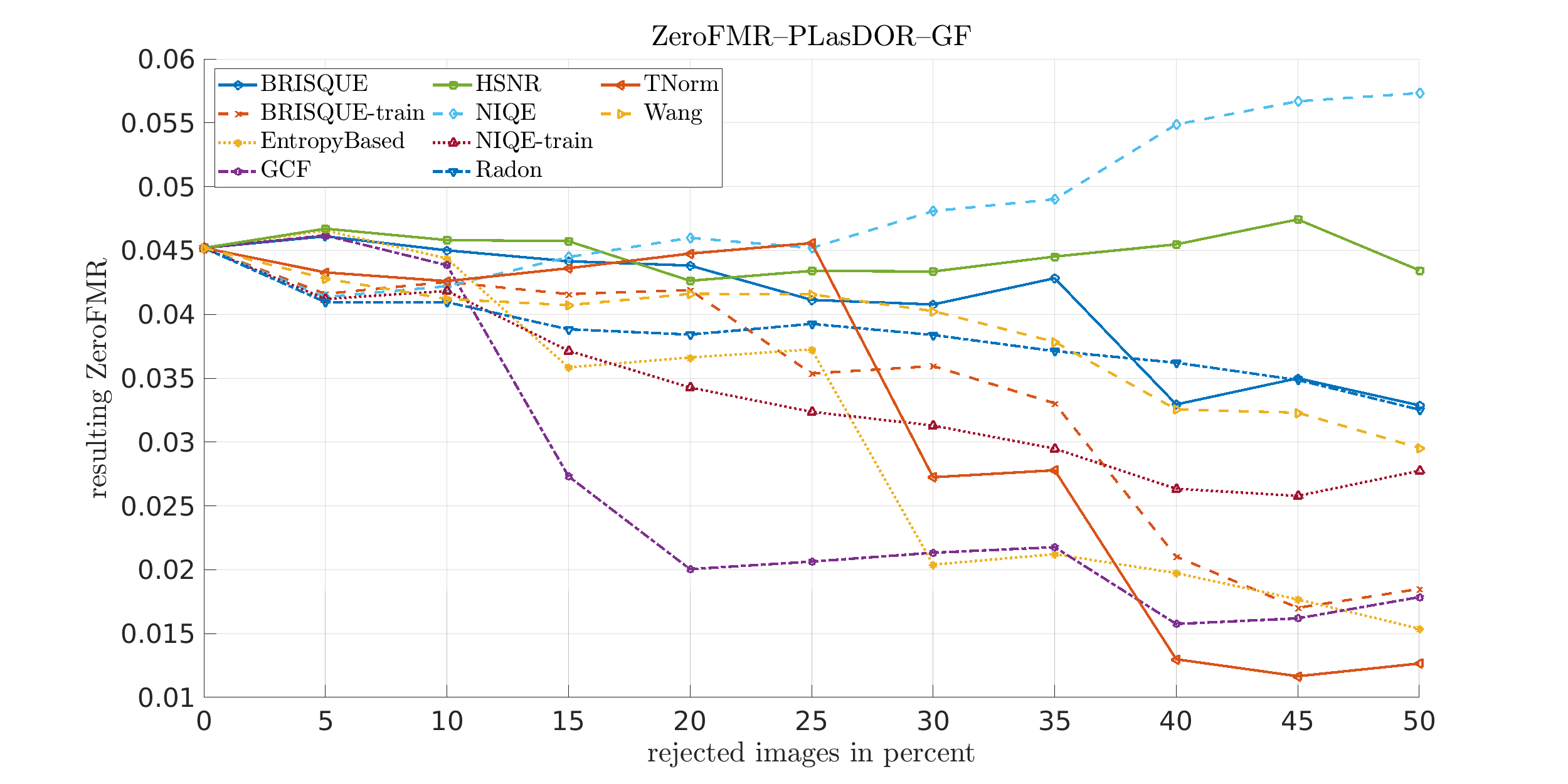}}
	\end{minipage}
	\begin{minipage}{0.33\textwidth}
	\centering{\includegraphics[width=1\textwidth]{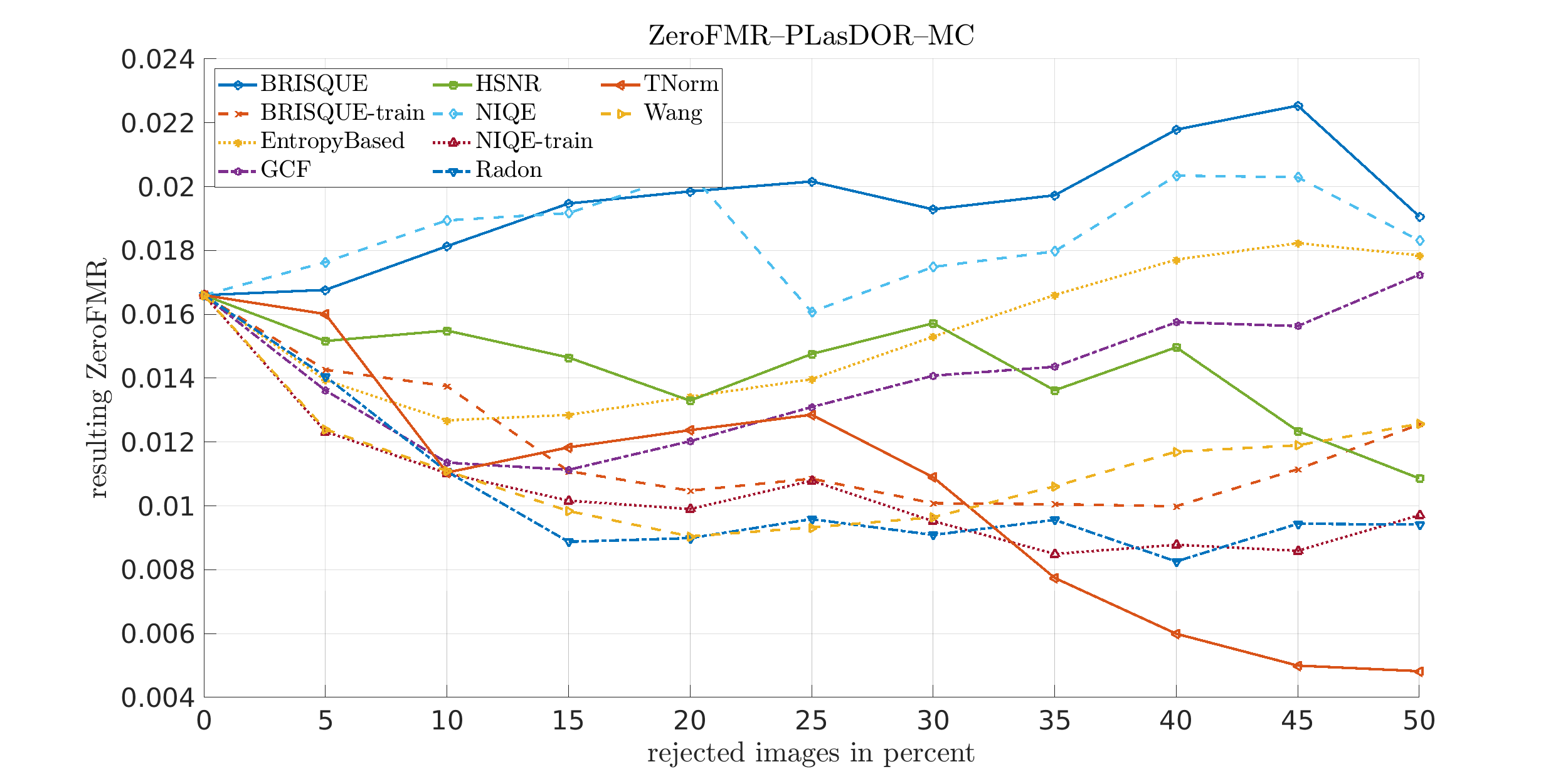}}
	\end{minipage}
	\begin{minipage}{0.33\textwidth}
	\centering{\includegraphics[width=1\textwidth]{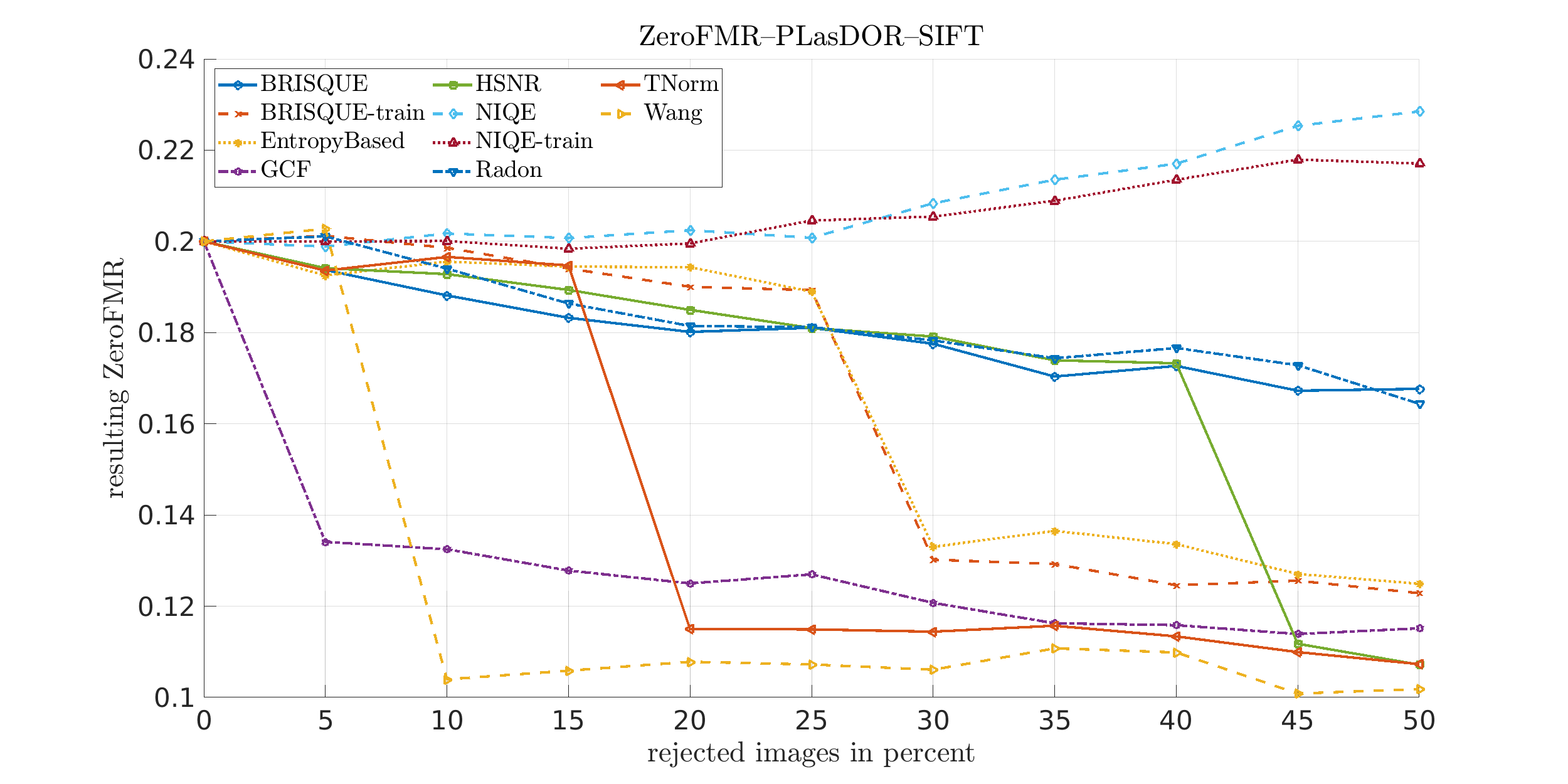}}
	\end{minipage} \vspace{0.5cm} \\ 
	\begin{minipage}{0.33\textwidth}
	\centering{\includegraphics[width=1\textwidth]{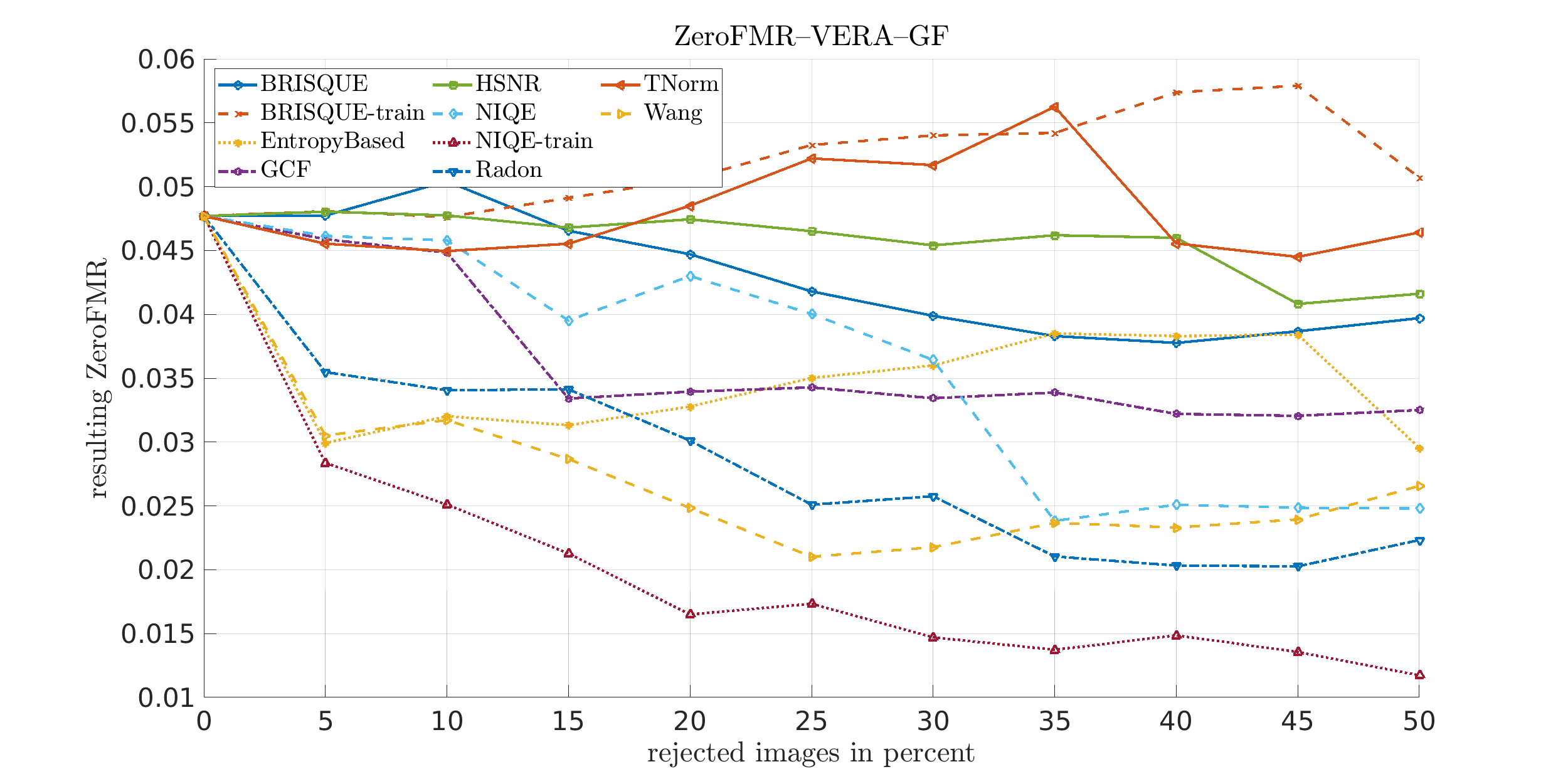}}
	\end{minipage} 
	\begin{minipage}{0.33\textwidth}
	\centering{\includegraphics[width=1\textwidth]{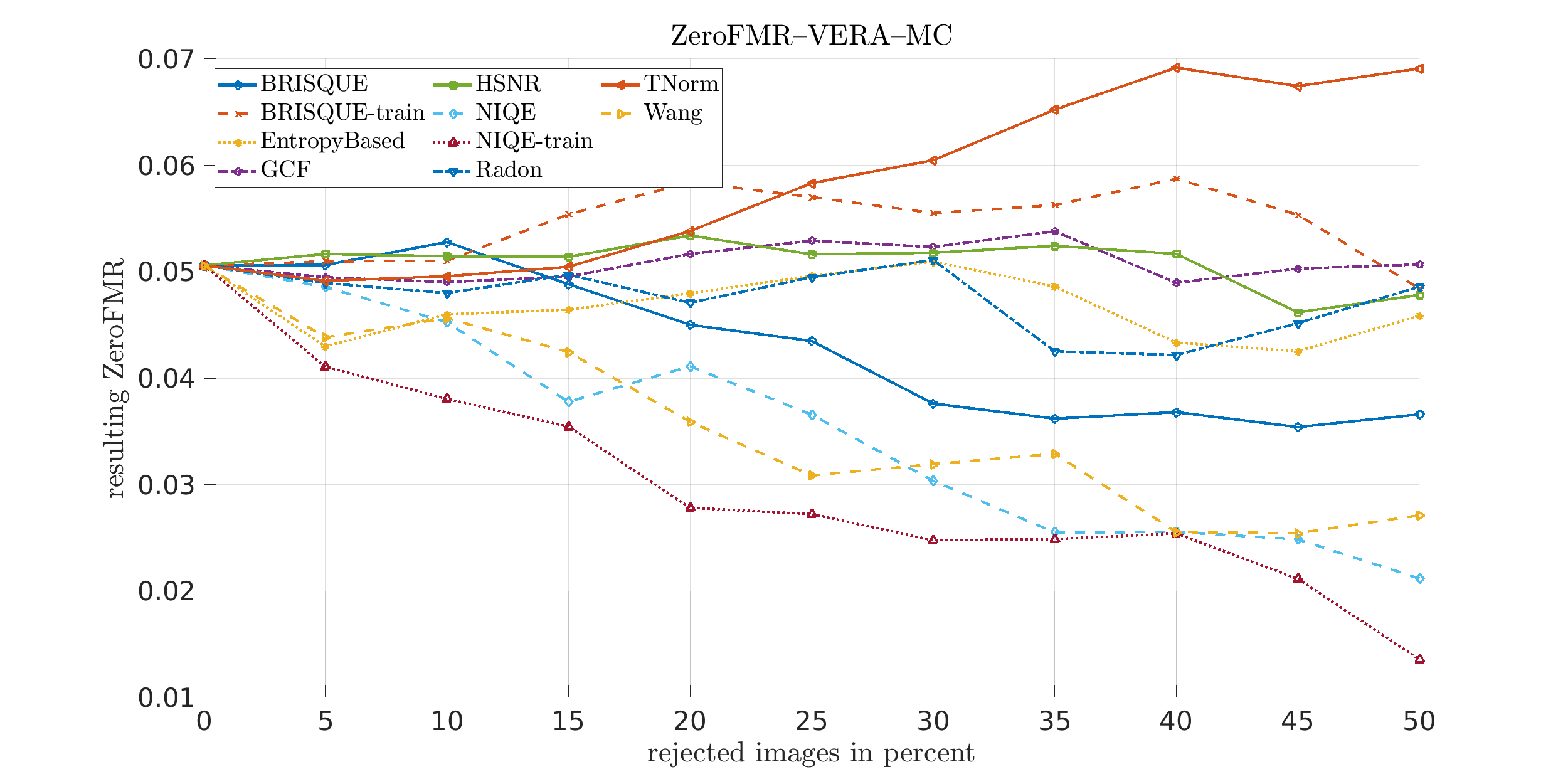}}
	\end{minipage} 
	\begin{minipage}{0.33\textwidth}
	\centering{\includegraphics[width=1\textwidth]{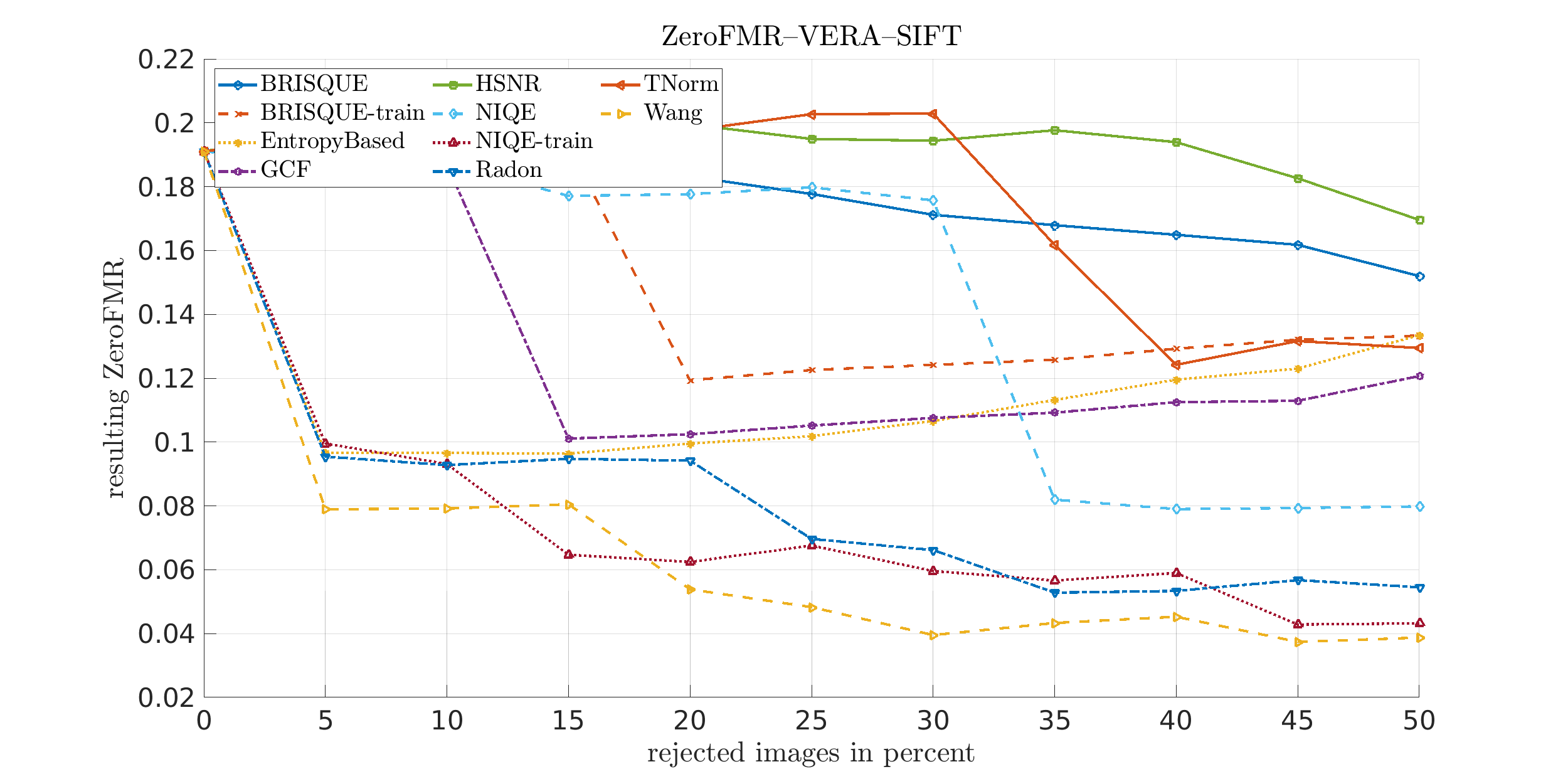}}
	\end{minipage} \\
	\caption{ZeroFMR trend with increasing rate of rejected images on PLasDOR (first row) and VERA (second row).}
	\label{fig: NEWvgl_Zero}
\end{figure*}

The following results focus on the extended evaluation based on the protocol of \cite{Remy22b} (third experiment) with the difference, that in this study a leave one database out evaluation is performed, whereas in \cite{Remy22b} samples of all databases were included during training. The leave one database out protocol was performed by selecting each of the databases listed in Section \ref{sec: datasets} using the metrics described in Sections \ref{sec: relatedWork} and \ref{sec: nss}. 
In this experiment an increasing number of low quality samples (according to the assessed quality scores) is discarded from the databases and not used during the sample comparisons. The samples are sorted according to the assigned quality scores and starting with excluding those $5\%$ images exhibiting lowest quality until excluding $50\%$. The recognition performance values (EER, FMR1000 and ZeroFMR) were re-computed using the remaining samples, resulting in the trend visualised in Figures \ref{fig: BIOSIGvgl}, \ref{fig: BIOSIGvgl_Zero}, \ref{fig: NEWvgl} and \ref{fig: NEWvgl_Zero}. For a well-performing and suitable sample quality evaluator, the EER should decrease with an increasing percentage of rejected images. From the EER progression two aspects can be observed immediately. First, the quality assessment application highly depends on the combination of database and feature extraction method. There is no general trend and thus, no conclusion can be drawn which quality evaluation method is suited best across all databases and feature types. Second, if only the EER is taken into account some of the quality measures do not work as expected, except for particular databases or feature extraction methods. For example, while Radon shows a poor performance on the PLasDOR database using GF, it turns out to be a good indicator for SIFT (both examples can be seen in the first row of Figure \ref{fig: NEWvgl}). In general, the obtained EER values stay within a low value range, which is a desirable observation from a recognition point of view, but a complicating factor regarding the significance of the EER based quality evaluation. 

TNorm exhibits a constant behaviour, an expectable monotonous decreasing EER, across all experiments focusing on finger vein images, except some outliers (e.g. see Figure \ref{fig: NEWvgl} top left plot). The same holds for Wang. NIQE-train is a suitable method for quality assessment, especially for VERA (see Figure \ref{fig: NEWvgl} right column) and CIE-HV as well as for most finger vein databases. Compared to BRISQUE-train NIQE-train is more stable and exhibits a monotonously decreasing EER in most cases, while the EER for BRISQUE-train is slightly increasing in several cases. The benefits of re-training NIQE are clearly visible as the original NIQE is usually among the worst performing metrics, clearly evident for UTFVP and MC (see middle plot of the upper row of Figure \ref{fig: BIOSIGvgl}). The results for VERA and CIE-HV are similar, thus the plots are omitted. For PTrans and PRefl only TNorm and Wang provide reasonably realistic values, for the other metrics, there is a clear, almost linear increase in the EER with an increasing number of discarded low quality images. Although fluctuations in the EER based analysis were to be expected, as can be seen in Figures \ref{fig: BIOSIGvgl} and \ref{fig: NEWvgl}, this can not be explained by differences in the visual quality of the hand vein database samples as no major changes can be detected (see Figure \ref{fig: HVimpressions}). \\
Compared to \cite{Remy22b}, the re-training of both BRISQUE and NIQE on UTFVP and SDUMLA does not improve the results as expected. As visualised in Figure \ref{fig: BIOSIGvgl} the EER is increasing with a high number of discarded low quality images. The most important difference between the current experiments and those in \cite{Remy22b} is the data used during the re-training of BRISQUE and NIQE. In \cite{Remy22b} a total of 50 poor and 50 good images each for UTFVP and SDUMLA have been utilised and no leave one dataset out training has been performed. In \cite{Remy22b} the focus was on the general robustness of the re-trained versions of BRISQUE and NIQE for a given biometric system where the capturing device as well the feature representation are known and thus, the quality assessment method can be optimised with respect to the particular biometric system. In the current study to focus is more on the generalisability of those metrics. In the current experiments all images from the selected databases, except the ones where it was evaluated on (UTFVP or SDUMLA for Figures \ref{fig: BIOSIGvgl} and \ref{fig: BIOSIGvgl_Zero}), are included in the training set. This updated training protocol was chosen to evaluate the generalisability of the re-trained methods. As the results reveal, the impact of including data from the same database during training or not is a crucial one. While in \cite{Remy22b} the re-trained methods show the expected trend of a decreasing EER with an increasing number of discarded samples, independent of the database and feature type, the current leave one database out evaluation results can not confirm this trend as described above. It is also worth noting, that compared to \cite{Remy22b} a parameter optimisation of the feature extraction methods was done to improve the baseline and overall recognition performance. This makes the existing subtle fluctuations in Figures\ref{fig: BIOSIGvgl} and \ref{fig: NEWvgl} more pronounced. \\
While the EER might not show the influence of a small number of low quality samples on the biometric recognition performance, especially for high number of samples, the ZeroFMR (FMR1000) is influenced even by a single low quality sample if it leads to a false negative comparison. Hence, the ZeroFMR, shown in Figures \ref{fig: BIOSIGvgl_Zero} and \ref{fig: NEWvgl_Zero}, allows a more detailed insight on the quality assessment performance. On the UTFVP and SDUMLA with a few exceptions (BRISQUE-train for UTFVP using all feature extraction methods, NIQE for UTFVP using GF and MC and NIQE for SDUMLA using SIFT) all of the evaluated quality estimators show the expected trend of a decreasing ZeroFMR towards a higher number of discarded low quality samples. On PLasDOR and VERA the same expected trend can be described, except NIQE for PLasDOR, BRISQUE for PLasDOR using MC, NIQE-train for PLasDor using SIFT as well as BRISQUE-train for VERA using GF and TNorm for VERA using MC. \\
Summing up, there is no best performing quality assessment methodology across all databases and feature types. All evaluated methodologies still show a high dependence of database and feature type. 

\begin{table}[ht!b] 
	\centering
	\caption{Overview on SVM classification accuracy over all selected categories (poor, middle, good) and databases (grouped into dorsal and palmar finger vein as well as hand vein).\label{tab: DLacc}}
	{\begin{tabular*}{23pc}{@{\extracolsep{\fill}}cccc@{}}\hline 
		folds & \textit{poor} & \textit{middle} & \textit{good}\\
		\hline  \vspace{0.2cm}
		\textbf{finger vein - dorsal} & \multicolumn{3}{l}{mean accuracy over all folds = 0.7250}\\
		mean accuracy per class & \textit{0.5594} & \textit{0.7461} & \textit{0.8088} \\
		\hline
		1-fold & 0.4193 & 0.7714 & 0.6558 \\
		2-fold & 0.7571 & 0.5641 & 0.8460 \\
		3-fold & 0.4260 & 0.7918 & 0.8322 \\
		4-fold & 0.6351 & 0.8571 & 0.9010 \\
		\hline\vspace{0.2cm}
		\textbf{finger vein - palmar} & \multicolumn{3}{l}{mean accuracy over all folds = 0.7336}\\ 
		mean accuracy per class & \textit{0.2404} & \textit{0.8348} & \textit{0.7820} \\
		\hline
		1-fold & 0.2705 & 0.8689 & 0.6960 \\
		2-fold & 0.2193 & 0.8495 & 0.6969 \\
		3-fold & 0.2003 & 0.8031 & 0.7425 \\
		4-fold & 0.2715 & 0.8179 & 0.7201 \\
		\hline\vspace{0.2cm}
		\textbf{hand vein} & \multicolumn{3}{l}{mean accuracy over all folds = 0.6773}\\ 
		mean accuracy per class & \textit{0.4894} & \textit{0.7447} & \textit{0.7224} \\
		\hline 
		1-fold & 0.5414 & 0.7507 & 0.6382 \\
		2-fold & 0.4196 & 0.7961 & 0.6779 \\
		3-fold & 0.4712 & 0.6901 & 0.7391 \\
		4-fold & 0.5251 & 0.7418 & 0.8344 \\
		\hline
\end{tabular*}}{}
\end{table}

\subsection{DL based vasculature quality assessment}
The last experiment evaluates the DL based quality measure. As mentioned in Section \ref{sec: expSetup} a 4-fold cross validation using a triplet loss function was done for dorsal finger vein data, palmar finger vein data and the remaining hand vein databases. The results, reflecting the classification accuracy after SVM application are presented in Table \ref{tab: DLacc} for each of the manually selected categories. All in all, the values presented in Table \ref{tab: DLacc} clearly indicate that for the hand vein and palmar finger vein data the accuracy for the middle class is highest, while for dorsal finger vein samples the highest accuracy is achieved for the good class. If only the mean accuracy over all folds for each of the three databases' groups is considered, the accuracy using palmar finger vein images is higher compared to the other two groups. However, this statement is only partly true. The classification accuracy of the manually categorised poor images is much lower compared to the middle and good category using palmar venous data and also lower compared to the accuracy of the poor class evaluating dorsal vein or hand vein images. There are two potential explanations for this observation. First, as described in Table \ref{tab: DBsplit}, the percentage of images contained in the poor class of the palmar finger vein databases is much lower compared to dorsal and hand vein ones. Furthermore, the difference between the number of vascular images exhibiting a poor manually assigned quality is much lower compared to the number of images labelled as middle or good (using palmar databases only). As a consequence, the portion of poor quality images selected during the training of the DL methods is low and thus, the proposed network is focusing (learning) more characteristics given for middle and good quality images, which makes it more difficult to assign poor data samples to the correct class. This statement is not only true for the palmar finger vein samples, the same trend (reduced classification accuracy of images exhibiting poor quality) is also found in the other two data groups, as the overall accuracy for classifying poor quality images correctly is also much lower for the other two groups. \\
Second, the differentiation between the single manually selected quality classes is not always an easy task (even for trained humans) due to the nature of the biometric trait. In general, the images exhibiting vasculature information are often of low contrast or illumination variations during the acquisition can lead to higher intensity values in some areas of the finger/hand (enabling a better visibility of the vein pattern), while other areas remain quite dark (reducing the likelihood of correctly determining the presence of vasculature patterns). For example such illumination variations can easily be detected by the trained network as problematic areas and thus, samples containing such areas were mistakenly classified as poor quality images (even if they are not). \\
Experiments reporting the EER performance as done for the other non DL quality techniques have not been conducted for the DL method. While for the non DL quality methods each image is labelled with a certain quality values (allowing to sort the databases and remove a certain portion of low quality samples), the proposed DL method only categorises each data sample into one of the three quality classes. Hence, a ascending or descending order can not be established and a removal of lower quality images can only be done on a randomised basis. As this procedure would not be comparable to the protocol used for the previous EER based analysis, this kind of evaluation was not performed.

\section{Conclusion}\label{sec: conclusion}
This work evaluated the suitability of several image quality assessment schemes as biometric quality estimators for vasculature pattern samples. In an extension of the previous work \cite{Remy22b}, additional databases as well as new DL based approach were included. Similar as in the previous work BRISQUE and NIQE turned out not to be suited as finger and hand vein quality measures if pre-trained on common images and classical distortions only. Their trained counterparts, re-trained on the vasculature pattern databases, exhibited a better performance. Although, compared to classical vasculature sample quality measures these re-trained versions of BRISQUE and NIQE do not necessarily perform better. The DL based approach achieved only mediocre classification performance and due to its limitation of three quality classes instead of a dedicated quality score was not satisfactory. In general, the results showed that the optimal quality measure highly depends on the selected database and feature representation, which is once again in-line with the previous findings in \cite{Remy22b}. \\
In the future, it is planned to refine and extend the DL based approach, enabling to output a dedicated quality score and improving its classification accuracy.

\section{Acknowledgments}\label{sec: acknowledgments}
This work has been partially supported by the Austrian Science Fund and Salzburg State Government, project no. FWF P32201.


\bibliographystyle{plain}
\bibliography{literature_journals, literature_conferences, literature_books, literature_standards}

\vfill\pagebreak

\end{document}